\RequirePackage{amsmath} 
\RequirePackage{fix-cm}
\documentclass[twocolumn]{svjour3}          
\smartqed  

\usepackage[table,dcipsnames]{xcolor}
\usepackage{graphicx}
\usepackage{comment}
\usepackage{microtype}
\usepackage{amssymb} 
\usepackage{color}

\usepackage[hyphens]{url}
\usepackage{color}
\usepackage{enumitem}
\usepackage[normalem]{ulem} 
\usepackage{framed}
\usepackage{tcolorbox}
\usepackage{wrapfig,lipsum}
\definecolor{EPIC-COLOR}{HTML}{ED323E}
\definecolor{EPIC-COLOR2}{HTML}{00D6D6}
\usepackage{multirow} 
\usepackage{ifthen} 
\usepackage{makecell} 
\definecolor{EPIC-BLUE}{HTML}{00B6D6} 
\definecolor{EPIC-RED}{HTML} {ED323E} 
\usepackage{booktabs} 
\usepackage{tabularx} 
\usepackage{pifont}  
\usepackage{siunitx} 
\usepackage{appendix}

\newcolumntype{R}{>{\raggedleft\arraybackslash}X}

\newcommand{\chParagraph}[1]{\noindent {\textbf{#1.}} \hspace{6pt}}

\newcommand{\edits}[1]{\textcolor{black}{#1}}

\newcommand {\oldDataset} {EPIC-KITCHENS-55{}}

\newcommand {\newDataset} {EPIC-KITCHENS-100}

\newcommand{\annStyle} {`pause-and-talk'{}}
\newcommand{\cmark}{\ding{51}}%
\newcommand{\xmark}{\ding{55}}%
\usepackage{xspace}
\newcommand*{\eg}{e.g.\@\xspace}
\newcommand*{\ie}{i.e.\@\xspace}

\journalname{IJCV}
\date{Received: 18 Jan 2021, Revised: 23 Aug 2021, Accepted: 17 Sep 2021}
\begin{document}
\emergencystretch 3em
\pagestyle{headings}

\title{\edits{Rescaling Egocentric Vision: Collection Pipeline and Challenges for EPIC-KITCHENS-100}} 
\authorrunning{D. Damen et al.}

\author{Dima Damen$^\dagger$ \and Hazel Doughty$^{\dagger \Diamond}$ \and
Giovanni Maria Farinella$^\ddagger$ \and
Antonino Furnari$^\ddagger$ \and
Evangelos Kazakos$^\dagger$ \and
Jian Ma$^\dagger$ \and
Davide Moltisanti$^{\dagger \star}$ \and
Jonathan Munro$^\dagger$ \and
Toby Perrett$^\dagger$ \and
Will Price$^\dagger$ \and
Michael Wray$^\dagger$}

\institute{$^\dagger$University of Bristol, UK \and $^\ddagger$University of Catania, Italy \and $^\Diamond$~Hazel is now at University of Amsterdam \and $^\star$ Davide is now at NTU Singapore
\\\textbf{Acknowledgements.} Research at Bristol is supported by Engineering and Physical Sciences Research Council (EPSRC) Doctoral Training Program (DTP), EPSRC Fellowship UMPIRE (EP/T004991/1). Research at Catania is sponsored by Piano della Ricerca 2016-2018 linea di Intervento 2 of DMI, by MISE - PON I\&C 2014-2020, ENIGMA project (CUP: B61B19000520008) and by MIUR AIM - Attrazione e Mobilita Internazionale Linea 1 - AIM1893589 - CUP E64118002540007. We thank David Fouhey and Dandan Shan from University of Michigan for providing the ego-trained hand-object detection model prior to its public release. We also thank Sanja Fidler from University of Toronto for contributing to the first edition of EPIC-KITCHENS. We appreciate the efforts of all voluntary participants to collect and narrate this dataset.}

\maketitle
\vspace{-2mm}
\begin{abstract}
This paper introduces the pipeline to \edits{extend} the largest dataset in egocentric vision\edits{,} EPIC-KITCHENS. The effort culminates in EPIC-KITCHENS-100, a collection of 100~hours, 20M frames, 90K actions in 700 variable-length videos, capturing long-term unscripted activities in 45 environments, using head-mounted cameras. 
Compared to its previous version~\cite{Damen2018EPICKITCHENS}, EPIC-KITCHENS-100 has been annotated using a novel pipeline that allows denser (54\% more actions per minute) and more complete annotations of fine-grained actions (+128\% more action segments).
This collection \edits{enables new challenges such as action detection and} evaluating the ``test of time'' --- \ie whether models trained on data collected in 2018 can generalise to new footage collected \edits{two years later}.

The dataset is aligned with 6 challenges: action recognition (full and weak supervision), action detection,  action anticipation, cross-modal retrieval~(from captions), as well as unsupervised domain adaptation for action recognition. For each challenge, we define the task, provide baselines and evaluation metrics\footnote{Dataset and leaderboards are available at \textcolor{blue}{\underline{\url{http://epic-kitchens.github.io/}}}}. 

\end{abstract}

\section{Introduction and Related Datasets}

Since the dawn of machine learning for computer vision, datasets have been curated to train models, for single tasks from classification~\cite{imagenet,carreira2017quo} to detection~\cite{coco,gu2018ava}, captioning~\cite{Karpathy14,xu2016msr} and segmentation~\cite{ade,Perazzi2016}. 
Increasingly, datasets have been used for novel tasks, through pre-training~\cite{he2018rip,Mettes2016shuffle}, self-supervision~\cite{noroozi2016,Vondrick2018tracking} or additional annotations~\cite{gupta2015visual,CabaJHG18}. However, task adaptation demonstrates that models overfit to the data and annotations~\cite{zhai2019largescale,Moltisanti2017}.

\begin{figure*}[t]
    \centering
    \includegraphics[width=\textwidth]{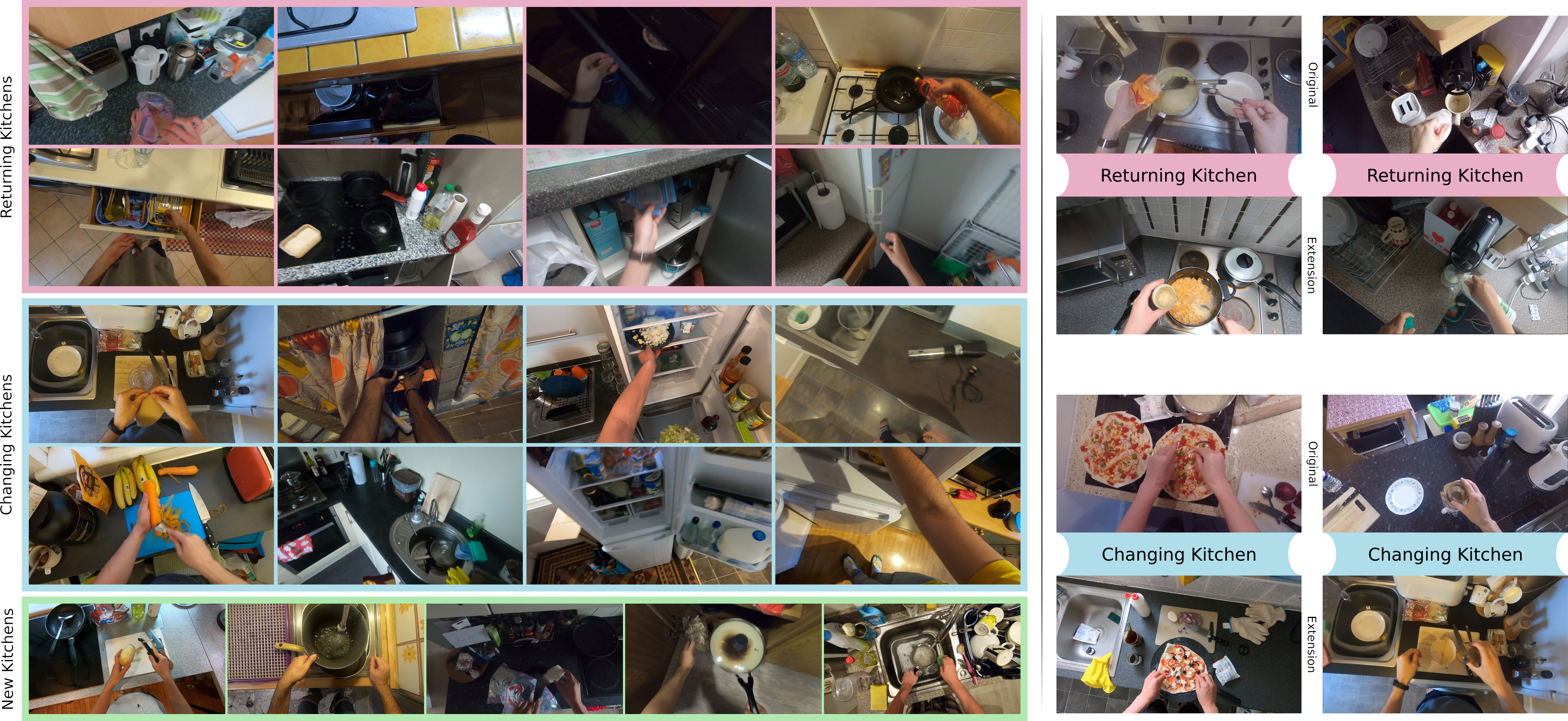}
    \caption{\edits{\textbf{Left:}} Frames from EPIC-KITCHENS-100 showcasing returning participants with returning or changing kitchens (top) as well as new participants (bottom). \edits{\textbf{Right:} Comparisons between recordings from~[1] and newly collected videos, with selected frames showcasing the same action. 
    Note object location differences in `returning' kitchens (e.g. microwave relocated). We show the same action performed in `changing' kitchens (e.g. same participant preparing pizza or filtered coffee in a new kitchen).}}
    \label{fig:kitchen_examples}
\end{figure*}

\begin{figure*}[htb]
    \centering
    \includegraphics[width=0.8\textwidth]{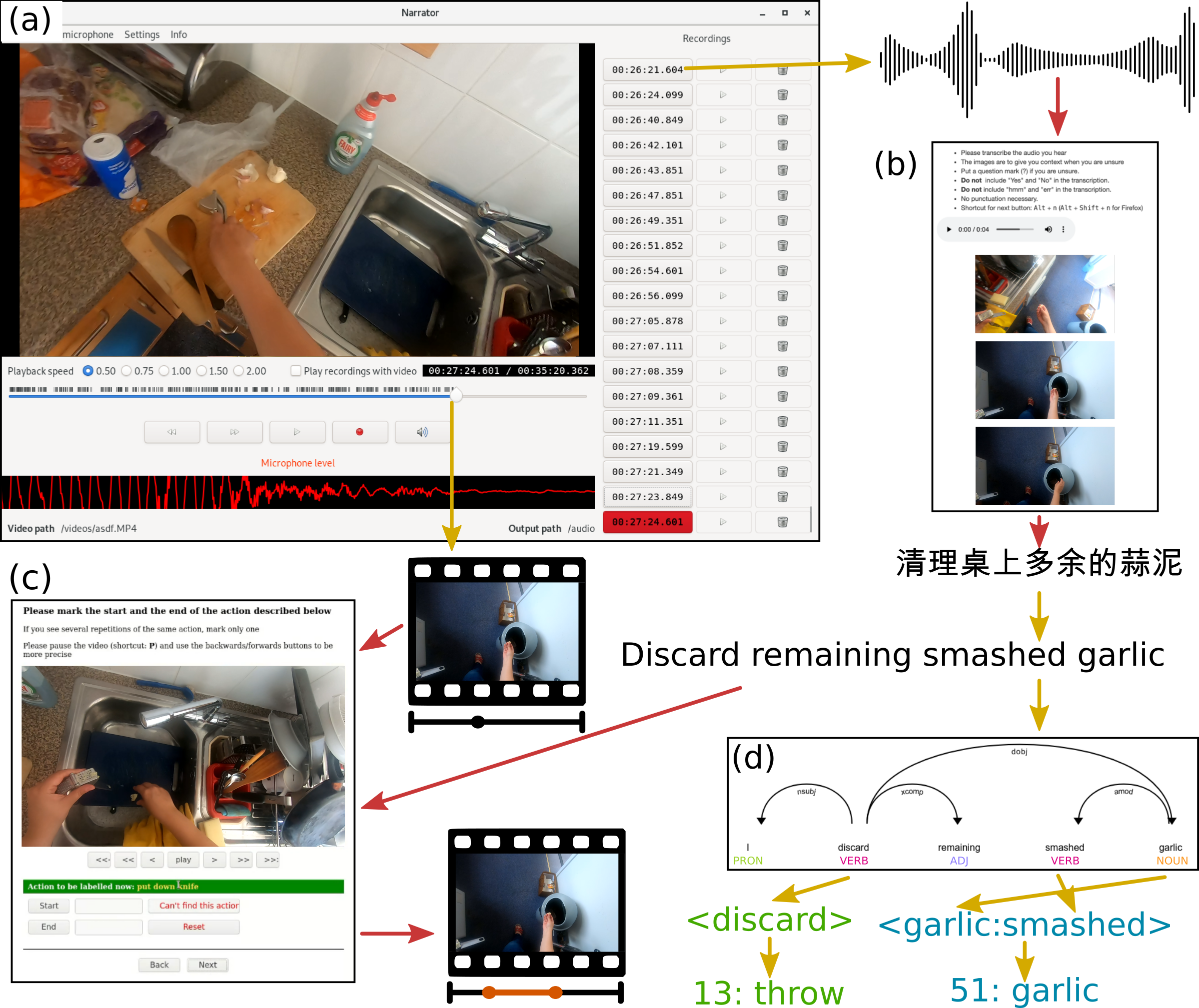}
    \caption{Annotation pipeline: (a)~narrator, (b) transcriber, (c) temporal segment annotator  and (d) dependency parser. Red arrows show AMT crowdsourcing of annotations.}
    \label{fig:main_annotation}
\end{figure*}

Alternatively, one dataset can be enriched with multiple annotations and tasks, \edits{aimed} towards learning intermediate representations through downstream and multi-task learning on the same input.  This has been recently achieved for autonomous driving~\cite{zhou2019does,geiger2012we,Cordts2016Cityscapes,neuhold2017mapillary,yu2018bdd100k,huang2018apolloscape,nuscenes2019,yogamani2019woodscape} and scene understanding~\cite{zamir2018taskonomy,silberman2012indoor}.
For example, 
Taskonomy~\cite{zamir2018taskonomy} contains 26 tasks ranging from edge detection to vanishing point estimation and scene classification. 

In comparison, the number of tasks proposed for \edits{action and activity understanding} datasets~\cite{Damen2018EPICKITCHENS,gu2018ava,caba2015activitynet,rohrbach2015dataset,zhou2017procnets,Rohrbach2012} remains modest.
Often, this is limited by the source of videos in these datasets. YouTube~\cite{caba2015activitynet,zhou2017procnets} and movies~\cite{gu2018ava,rohrbach2015dataset} typically contain curated videos, with edited shots.
However, attempts to define multiple challenges for these datasets have been exemplary. ActivityNet~\cite{caba2015activitynet} is the most popular video challenge, evaluated for localisation, dense captioning~\cite{krishna2017dense} and object detection~\cite{zhou2019grounded}.
Similarly, AVA~\cite{gu2018ava} has  challenges on action localisation and active speaker detection~\cite{roth2019ava}.

Several leading egocentric datasets~~\cite{Pirsiavash2012,Damen2014a,Fathi2012,de2008guide,EGTEA} showcased the unique perspective and potential of first-person views for action recognition, particularly hand-object interactions.
In 2018, the introduction of the largest-scale dataset EPIC-KITCHENS~\cite{Damen2018EPICKITCHENS}\ has transformed egocentric vision, not only due to its size, but also the unscripted nature of its collection and the scalable nature of the collection pipeline.
In this paper, we present \newDataset, a substantial extension which brings the total footage to \textbf{100} hours, capturing diverse unscripted and unedited object interactions in people's kitchens\footnote{We will refer to the previous edition as EPIC-KITCHENS-55 in reference to the number of hours collected and annotated.}. 
\edits{As shown in Fig 1, the actions capture hand object interactions with everyday objects in participants' kitchens. The unscripted nature of the dataset results in naturally unbalanced data, with novel compositions of actions in new environments. While challenging, the dataset is domain-specific (i.e. kitchen-based activities), offering opportunities for engaging domain knowledge. We offer two-level annotations for nouns and verbs in interactions (e.g. ``carrot/courgette $\rightarrow$ vegetable'', ``put/throw/drop $\rightarrow$ leave'') to utilise such priors.}

\edits{Importantly, we propose a refined} annotation pipeline that results in denser and more complete annotations of actions \edits{in untrimmed videos.}
This pipeline enables various tasks on the same dataset; we demonstrate six in Section \ref{sec:challenges}, with baselines and evaluation metrics that
focus on understanding fine-grained actions and offer benchmarks which can support research into better modelling of video data.

\section{Data Collection and Scalable Pipeline}
\label{sec:data_collection_pipeline}
In this section, we detail our collection and annotation effort. 

\chParagraph{Data Collection}
We collect additional footage as follows:
we contacted participants from \oldDataset{} to record further footage.
Of the 32 participants in~\cite{Damen2018EPICKITCHENS}, 16 subjects expressed interest in participating. Interestingly, half of these (8 subjects) had moved homes over the past two years.
We also recruited 5 additional subjects, increasing the total number of subjects and kitchen environments to 37 and 45 respectively.
All participants were asked to collect 2--4 days of their typical kitchen activities, as in~\cite{Damen2018EPICKITCHENS}.
We collect footage using a head mounted GoPro Hero7 black. This is two generations newer than the camera used in~\oldDataset, with a built-in feature for HyperSmooth video stabilisation. 
Sample frames are shown in Fig. \ref{fig:kitchen_examples}, with selected frames of the same action in returning and changing kitchens.

\chParagraph{Annotation Pipeline} An overview of the pipeline can be seen in Fig. \ref{fig:main_annotation}.

\chParagraph{(a) Narrator}
Previously, for~\oldDataset, we used a \edits{non-stop} narration approach, where 
\edits{each participant narrated their previous action while watching the future actions in the running video. We found this resulted in increased mental load and some}
actions being missed or misspoken.
To improve upon this approach, we take inspiration from~\cite{Gigli2019}, where
objects in images are annotated by pointing and speaking and propose temporal `pointing' which we refer to as \annStyle.
By allowing participants to pause the video to speak  \edits{as well as take breaks}, we hope to increase accuracy and density of actions, whilst still allowing for a scalable narration approach.
We built an interface to facilitate collecting such narrations from the participants (Fig. \ref{fig:main_annotation}a), which includes a video player, synced with audio recordings\footnote{Our tool is available at \resizebox{\linewidth}{!}{\textcolor{blue}{\underline{\url{https://github.com/epic-kitchens/epic-kitchens-100-narrator}}}}}. Participants watch the video and press a key to pause while they narrate the action in their native language.
As previously observed in~\cite{Damen2018EPICKITCHENS}, using the native language ensures the narrations use the correct vocabulary in describing the actions.
The video restarts on key release.
\edits{Note that the narrator still watches the video \emph{once}, maintaining the targeted scalability of the annotation pipeline, but removes the mental overload of narrating past actions while watching future actions.}
This allows for short and overlapping actions to be captured in addition to enabling error correction, as participants can listen to, delete or re-record a narration.
Fig. \ref{fig:main_annotation} shows an ongoing narration, demonstrating density (ticks on the slider). 

\begin{figure*}[t!]
    \centering
    \includegraphics[width=\textwidth]{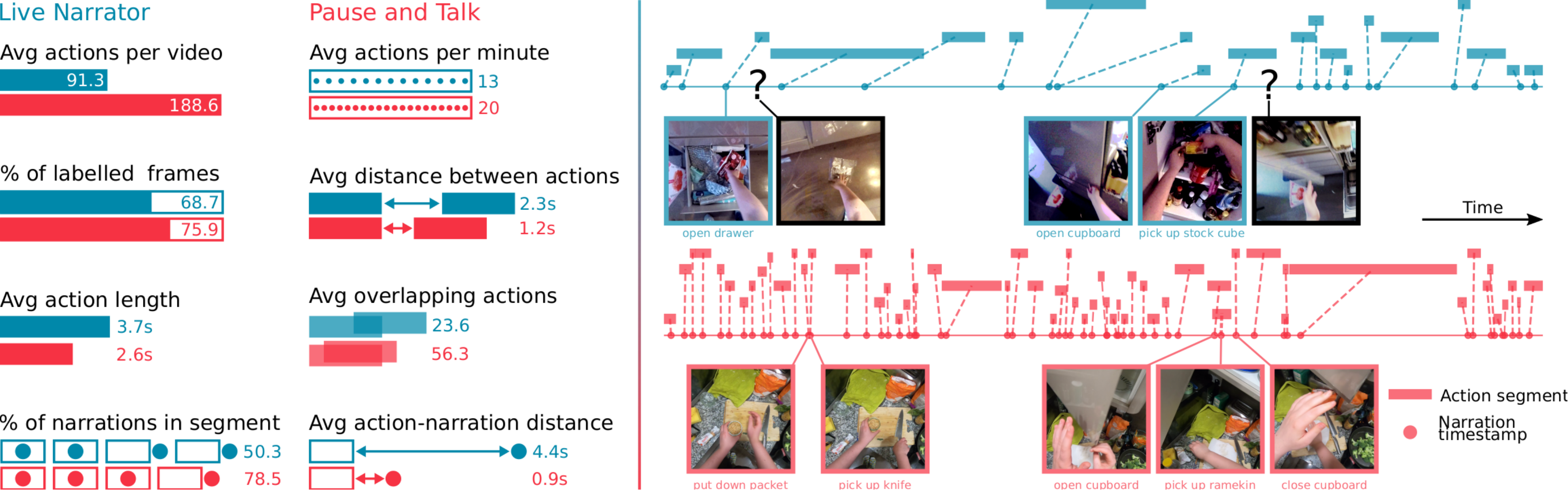}
    \caption{Comparing \edits{non-stop} narrations (blue) to \annStyle{} narrations (red). Right: timestamps (dots) and segments (bars) for two sample sequences. \annStyle{} captures all actions including short ones. Black frames depict missed actions.}
    \label{fig:qual_comp_18_20}
\end{figure*}

\chParagraph{(b) Transcriber}
We perform transcription of audio narrations, followed by translation (if applicable): first, we transcribe all narrations and then translate the unique transcriptions into English using a hired translator for correctness and consistency.
The approach we used to transcribe narrations in~\cite{Damen2018EPICKITCHENS} had issues where workers failed to understand some audio narrations due to the lack of any visual information.
To mitigate this, we build a new transcriber interface containing three images sampled around the timestamp (Fig. \ref{fig:main_annotation}b).
We find that images increase worker agreement and alleviate issues with homonyms (\eg~`flower' and `flour').
Each narration is transcribed into a caption by 3 Amazon Mechanical Turk (AMT) workers using a consensus of 2 or more workers.
Transcriptions were automatically rejected if the cosine similarity between the Word2Vec~\cite{mikolov2013efficient} embeddings was lower than an empirical threshold of 0.9.
When AMT workers fail to agree, the correct transcription was selected manually.
Captions were then spell checked and substitutions were applied from a curated list of problematic words (\eg `hob' and `hop'), further reducing errors.

\chParagraph{(c) Parser}
We use spaCy~\cite{spacy2} to parse the transcribed captions into verbs and nouns (Fig. \ref{fig:main_annotation}c) and manually group these into minimally overlapping classes as we did in our previous work.
We reworked this to improve parsing of compound nouns and missing verbs/nouns. Additionally, all annotations (including those we collected previously from \oldDataset) were re-parsed using the updated pipeline.
To cluster the verbs and nouns, we adjust previous clusters to reduce ambiguities between classes. For example, we group `brush' and `sweep' into one verb class, and introduce noun classes that did not exist before such as `lentils'.

\chParagraph{(d) Temporal Annotator}
We built an AMT interface for labelling start/end times of action segments (Fig. \ref{fig:main_annotation}d).
Annotators completed a quick tutorial on annotating temporal bounds before they labelled 10 consecutive actions.
To create the bounds of the action segment, we use the same approach as we did previously but increased the number of workers from 4 to 5 to improve quality.
\edits{Note that in the untrimmed videos there might be consecutive instances of the same action. These will be indicated by repeated narrations. We thus request that annotators mark the temporal bounds of each instance, prompted by the timestamp. This avoids the merging of instances of the same action.}

\chParagraph{Quality Improvements}
Our \newDataset{} scalable pipeline focuses on denser and more accurate annotations. We compare different parts of the pipeline to our previous one in Appendix~\ref{app:sectionB}.
Here, we show improved quality of annotations both numerically and through an example.

Fig. \ref{fig:qual_comp_18_20} (left) compares the narration method we used in~\cite{Damen2018EPICKITCHENS} to the new pipeline over several metrics. Our \annStyle{} narrator produces more densely annotated videos; fewer gaps and more labelled frames; actions are shorter; and exhibit higher overlap. 
The narration timestamps are also closer to the relevant action, with a higher percentage being contained within the action and a smaller distance to remaining timestamps outside the action.

Fig. \ref{fig:qual_comp_18_20} (right) shows two video sections, of equal length, annotated by the same participant, one using \edits{non-stop} narrations and the other with \annStyle.
The number of annotated actions increased from 20 to 56, with short actions (such as `turn on tap') often missed in the previous pipeline.
We demonstrate these through two examples.
The first shows a missed action of  picking up a bag off the floor that had been dropped, and the second shows a missed closing cupboard action.
In the sequence from \annStyle{}, all actions including closing the cupboard were successfully narrated thanks to our \annStyle{} pipeline. By narrating more actions, the start/end times also become more accurate as it is more obvious to the AMT annotators what each narration refers to.

\section{Statistics, Scalability and the Test of Time}
\label{sec:statistics}

\begin{table*}[t]
    \centering
        \caption{Statistics of \newDataset{} and its Train/Val/Test splits. \edits{*There is overlap between the unique narrations of~\cite{Damen2018EPICKITCHENS} and extension, hence overall does not sum up across sources.}}
\resizebox{\textwidth}{!}{%
    \begin{tabular}{ll
        S[table-format=2.1]
        S[table-format=3.0]
        *{2}{S[table-format=5.0]}
        S[table-format=2.0]
        S[table-format=3.0]
        S[table-format=4.0]
        *{3}{S[table-format=8.0]}
      }
      \toprule
        &                                          & {Hours} & {Videos} & {Action Seg.} & {Unique Narr.} & {Verb Cls.} & {Noun Cls.} & {Action Cls.} & {Object Masks} & {Hand BB} & {Int. Obj} \\ \midrule
        \multirow{3}{*}{\rotatebox{90}{\textbf{Source}}} & Videos from~\cite{Damen2018EPICKITCHENS} & 54.6 & 432 & 39,432 & 11,423 & 93 & 272 & 2,747 & 35,682,398 & 18,234,678 & 22,156,746\\
        & Extension                                & 45.4 & 268 & 50,547 & 11,236 & 91 & 266 & 2,900 & 29,987,598 & 12,999,913 & 16,043,057\\ 
        &\bfseries Overall                         &\bfseries 100.0 & \bfseries 700 & \bfseries 89,977 & \bfseries 20,580* & \bfseries 97 & \bfseries 300 & \bfseries 4,053 & \bfseries 65,669,996 &\bfseries 31,234,591 &\bfseries 38,199,803
 \\ \midrule
        
        \multirow{3}{*}{\rotatebox{90}{\textbf{Splits}}} & Train                                    & 74.7 & 495 & 67,217 & 15,968 & 97 & 289 & 3,568 & 48,896,723 & 23,186,294 & 28,190,446\\
        & Val                                      & 13.2 & 138 &  9,668 &  3,835 & 78 & 211 & 1,352 &  8,714,871 & 4,462,472  & 5,513,884\\
        & Test                                     & 12.1 &  67 & 13,092 &  4,324 & 84 & 207 & 1,487 &  8,058,402 & 3,585,825  & 4,495,473\\
         \bottomrule
    \end{tabular}%
    }
    \label{tab:dataset_split_stats}
\end{table*}
\begin{figure*}[t]
    \centering
    \includegraphics[width=\textwidth]{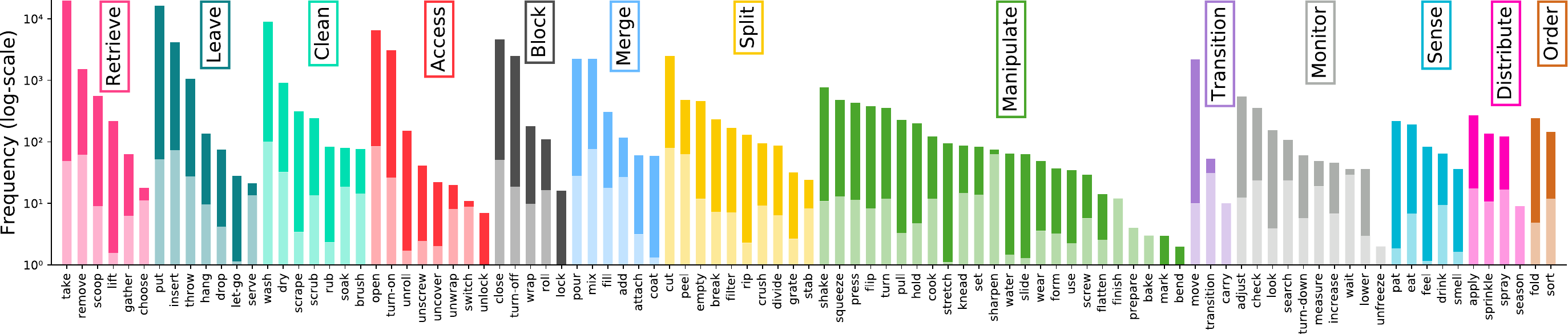}
    \includegraphics[width=\textwidth]{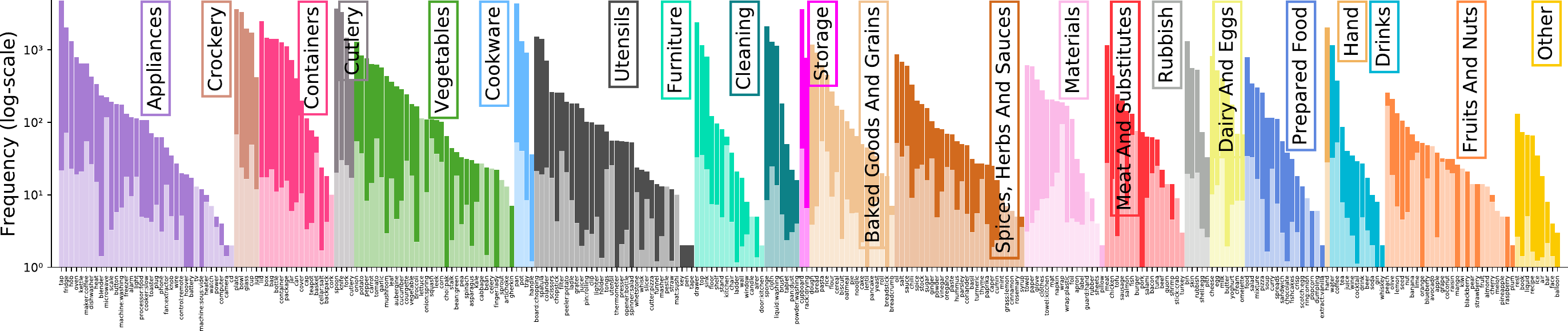}
    \caption{Frequency of verbs (top) and nouns (bottom), grouped by category. Each bar is \textit{linearly} split: solid represents instances from newly-collected videos and washed-out from original videos.}
    \label{fig:verb_noun_categories}
    \includegraphics[width=\textwidth]{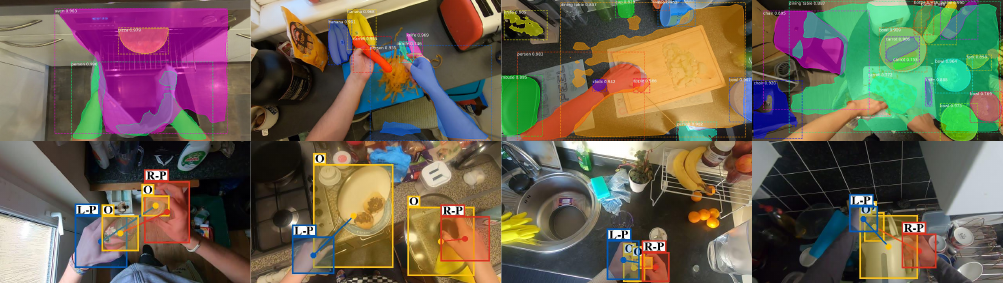}
    \caption{\textbf{Top:} Sample Mask R-CNN of large objects (col1: oven), hands (labelled person), smaller objects (col2: knife, carrot, banana, col3: clock, toaster, col4: bottle, bowl), incorrect labels of visually ambiguous objects (col3: apple vs onion) and incorrect labels (col3: mouse, col4: chair).
    \textbf{Bottom:} Sample hand-object detections from \cite{Shan2020Understanding}. \textit{L/R} = Left/Right, \textit{P} = interaction with portable object, \textit{O} = object. Multiple object interactions are detected (col2: pan and lid, col4: tap and kettle).}
    \label{fig:masks}
\end{figure*}

\begin{figure*}[t]
    \begin{center}
    \includegraphics[width=1\linewidth]{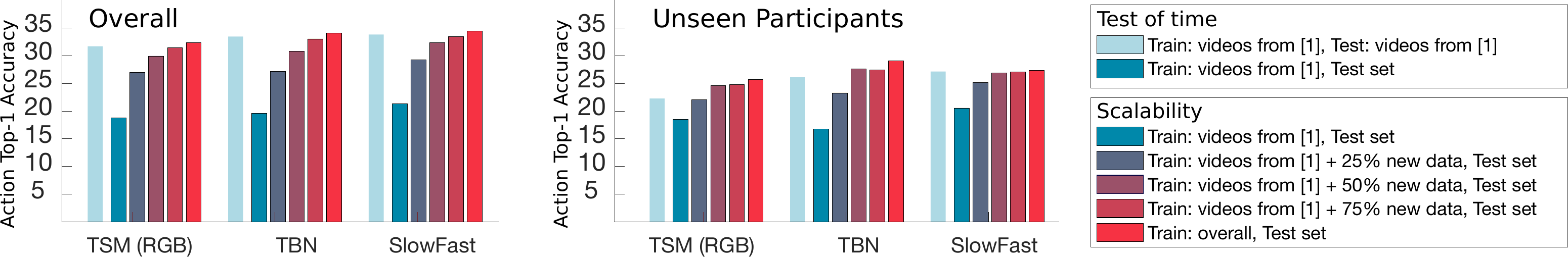}
    \caption{Test of time and scalability test results.}
    \label{fig:scalability}
    \end{center}
\end{figure*}

\newDataset{} contains 89,977 segments of fine-grained actions annotated from 700 long videos. 
Footage length amounts to 100 hours. 
Table \ref{tab:dataset_split_stats} lists the general statistics, separating those from the videos collected previously to the newly collected videos. Note that all previous narrations have been re-parsed using the new pipeline (Fig.~\ref{fig:main_annotation}b-d).
\newDataset{} rescales our previous dataset with almost double the length with $1.8$x hours and $2.3$x action segments. Comparisons to other datasets are presented under relevant benchmarks in Section~\ref{sec:challenges}.

In Fig.~\ref{fig:verb_noun_categories} we show the frequency of verb (97) and noun (300) classes in the dataset.
These are grouped into categories (13 verb and 21 noun categories), sorted by size. 
For example, verbs `wash', `dry', `scrape', `scrub', `rub', `soak' and `brush' are grouped into a \textit{clean} verb category. 
The plots show a clear long-tail distribution.
The contribution of each class from source videos~\cite{Damen2018EPICKITCHENS} and extension are also shown. 
New verb classes (\eg `lock', `bend') and noun classes (\eg `orange' and `hoover') are only present in the newly-collected videos.

We enrich our dataset with automatic spatial annotations using two models. The first is Mask R-CNN~\cite{he2017mask} trained on MSCOCO~\cite{coco}. The second is hand-object interactions from~\cite{Shan2020Understanding}, trained on 100K images from YouTube along with 42K images from three egocentric datasets~\cite{Damen2018EPICKITCHENS,sigurdsson2018charadesego,EGTEA} of which 18K are from our videos~\cite{Damen2018EPICKITCHENS}. It detects interacted static and portable objects as an offset to hand detections.
Example annotations are shown in
Fig.~\ref{fig:masks}, and the number of annotations is given in Table~\ref{tab:dataset_split_stats}.
While we do not use these annotations to report results, we believe these 66M masks, 31M hand and 38M object bounding boxes could facilitate future models for spatial (or spatio-temporal) attention\footnote{Correctness of bounding boxes for hands and objects has been evaluated by Shan et al.~\cite{Shan2020Understanding} - see acknowledgements. Performance of R-CNN masks has not been quantitatively evaluated and these are error-prone.}.

\chParagraph{Splits}
We split the videos into Train/Val/Test with a ratio of roughly 75/10/15. Each video, with all its action segments, is present in one of the splits, and the  
Test split contains only newly-collected videos.
We use re-parsed videos from the original EPIC-KITCHENS test sets\footnote{We no longer split the test set into seen and unseen kitchens, but instead report on relevant evaluation metrics for each challenge.} as the new validation set.
Our Val/Test splits contain two interesting subsets, which we report on separately:
\begin{itemize}[leftmargin=*,itemsep=-2ex,partopsep=1ex,parsep=2ex]
    \item \emph{Unseen Participants}: Our Val and Test splits contain participants not present in Train: 2 participants in Val, and another 3 participants in Test. These contain 1,065 and 4,110 action segments respectively.
    This subset helps evaluate the generalisability of the models across the various benchmarks.
    \item \emph{Tail Classes}: We define these (for verbs and nouns) to be the set of smallest classes whose instances account for 20\% of the total number of instances in training.
A tail action class contains either a tail verb class or a tail noun class.
These are 86/228/3,729 verb/noun/action classes. 

\end{itemize}

\chParagraph{Scalability and the Test of Time}
As we rescale EPIC-KITCHENS with additional videos, we carry out two investigations: (a) how models trained on videos from~\cite{Damen2018EPICKITCHENS} perform on videos collected two years later, and (b) how  models' performance scales with additional annotated data. We call these the \textit{test of time} and the \textit{scalability} tests respectively.

Fig.~\ref{fig:scalability} includes results for both investigations, evaluated on the task of action recognition (definition and models from Section~\ref{sec:action_recognition_challenge}).
We separate overall results~(left) from unseen participants (right). For all models, comparing the first two bars demonstrates that models trained solely on videos from~\cite{Damen2018EPICKITCHENS} do not withstand the \textit{test of time}. For the same model, performance drops significantly when new data is evaluated. This highlights a potential domain gap, which we discuss next.
We assess \textit{scalability}
 by gradually adding new data in training. Results demonstrate a significant improvement, albeit saturating when 50\% of new data is added, particularly for unseen participants. This highlights the need for better models \edits{and more diverse data rather than merely more data. This can be particularly observed as the unseen participants data benefits even less when scaling. We tackle the gap to new environments and participants next.}

\chParagraph{Unravelling the Domain Gap}
As defined in the early work on speech recognition
\cite{Ueberla1997},
``A domain $D$ is a (often infinite) set of samples such that each sample satisfies a property $P_D$''.
A domain gap is present when at least one property differs between the samples of two domains.
Domain gaps have been a frequent source of frustration for a wide range of learning tasks, where models are trained on samples from one domain, and thus under-perform when deployed in a different domain. 
This is also known as sample-selection bias \cite{Heckman1979}.
\edits{Sampling bias is a common cause for a domain gap between datasets, which can not easily be removed during dataset collection, as noted in~\cite{Torralba2011}.}
The most obvious domain gaps stem from changes in locations \cite{Oberdiek2020}, viewpoints~\cite{Zhai2017}, labels~\cite{Hsu2020} and participants \cite{Stein}. However, there are often more subtle causes, such as differences in capture methodology \cite{Saenko2010a}
or due to \edits{changes in objects, environments and actions over time.}

The concept of a compound domain gap has recently been introduced in \cite{Liu2020}, where the target domain is a compound of multiple domains without domain labels. As stated by Liu et al.~\cite{Liu2020}, this is a more realistic scenario resulting from unconstrained data collection. 
In \newDataset{}, each video in the extension offers a compound domain gap due to changes in one or more of the following properties: 
\begin{itemize}[leftmargin=*,itemsep=-2ex,partopsep=1ex,parsep=2ex]
    \item Hardware and capturing as in \cite{Saenko2010a,Gong2012}.  Extended footage uses a newer camera model with onboard video stabilisation.
    \item Locations as in \cite{Oberdiek2020}.  As indicated in Section~\ref{sec:data_collection_pipeline},  eight subjects have moved home resulting in changing surroundings but keeping the appearance of many objects and tools. Additionally, unseen participants capture footage in new environments where the appearance of objects and surroundings differ.
    \item Participants as in \cite{Stein}.  Hand appearance and individual behaviours exist in the extension which are not in the original footage.
    \item Short-term temporal offsets as in \cite{Wulfmeier2018}, where time-of-day can affect scene lighting, and some background objects change position (e.g. on the counter for one video, put away in a cupboard for a later video).
    \item Long-term temporal offsets as in \cite{Carlevaris-Bianco2016,Maddern2017}.  \newDataset{} is filmed 2 years after \oldDataset{}.  In the same environment, changes such as wear and tear, new objects and different object positions are observed (see Fig~\ref{fig:kitchen_examples} right).  Participant behaviour can also change over time.
\end{itemize}

While we have domain labels for some of these properties (e.g. recording camera, location, time-of-day and participant ID), other property changes can vary between samples, without associated labels.
It is particularly difficult to associate labels with changes in behaviour or object appearances, for example.
We publish these properties with the dataset when present.
Importantly, we explore this compound domain gap, without using property labels, using a new challenge on unsupervised adaptation for action recognition (Section~\ref{sec:domain_adaptation_challenge}).

\section{Challenges and Baselines}
\label{sec:challenges}
In this section, we define 6 challenges on our dataset, two modified from~\cite{Damen2018EPICKITCHENS}, namely action recognition (Section~\ref{sec:action_recognition_challenge}) and anticipation (Section~\ref{sec:action_anticipation_challenge}). We introduce four new challenges: weakly-supervised action recognition (Section~\ref{sec:action_weakly_recognition_challenge}), action detection (Section~\ref{sec:action_detection_challenge}),  unsupervised domain adaptation for action recognition~(Section~\ref{sec:domain_adaptation_challenge}) and action retrieval (Section~\ref{sec:action_retrieval_challenge}). 
While many works have addressed one or more of these challenges, they are typically explored using different datasets. Our annotation pipeline (from captions and single timestamps to segments and classes---Fig.~\ref{fig:main_annotation}) can be used to define multiple challenges, potentially jointly. In this section, we only scratch the surface by reporting on each challenge independently. For readability, we include all implementation details in Appendix~\ref{sec:app_impl_details}, and we published all our baseline models and evaluation scripts.
\subsection{Action Recognition}
\label{sec:action_recognition_challenge}
\chParagraph{Definition} As in~\cite{Damen2018EPICKITCHENS}, we consider a video segment ($t_s, t_e$) as the start and end frames in a video. We aim to predict ($\hat{v}, \hat{n}, \hat{a}$) as the verb/noun/action classes of the action in this segment. We consider overlapping segments independently.

\begin{table*}[t!]
  \centering
  \caption{A comparison of \newDataset{} against popular action recognition datasets. \mbox{\textit{a} = Action,} \textit{v} = Verb, \textit{n} = Noun, \textit{c} = caption, \textit{ML-a} = Multi-Label Action.}

\label{tab:ar_comparison}
  \resizebox{\textwidth}{!}{%
\begin{tabular}{@{}lr@{\enskip}l@{\enskip}r@{\enskip}r@{\enskip}r@{\enskip}r@{\enskip}r@{\enskip}l@{\enskip}l@{}}
\toprule
                    \textbf{Dataset}
                    & \thead{Year}
                    & \thead{Type}
                    & \thead{Hours}
                    & \thead{Actions\\per Video}
                    & \thead{Action\\Clips}
                    & \thead{Avg. Action\\Length}
                    & \thead{Action\\Classes}
                    & \thead{Task}
                    & \thead{Metrics}      \\ \midrule
\newDataset{}                                    & 2020 & Unscripted & 100  & 128.5 & 90k  & $3.1\pm5.4$  & 4,053 & a, v, n & Top-1/5 Acc. \\
Kinetics-700~\cite{Kay2017,kinetics700}                      & 2019 & YouTube    & 1806 & 1     & 650k & $10.0\pm0.0$ & 700   & a       & Top-1/5 Err. \\
Multi-Moments in Time~\cite{monfort2020moments}  & 2019 & YouTube    & 850  & 1     & 1M   & $3.0\pm0.0$  & 313   & ML-a    & mAP          \\
Something-Something V2~\cite{Goyal2017,mahdisoltani2018effectiveness}             & 2018 & Scripted   & 234  & 1     & 220k & $3.8\pm1.1$  & 174   & a, n, c & Top-1/5 Acc. \\
AVA~\cite{roth2019ava}                           & 2018 & Film       & 108  & 1380  & 410k & $2.7\pm3.5$  & 80    & ML-a    & mAP          \\
HACS~\cite{zhao2019hacs}                         & 2017 & YouTube    & 861  & 2     & 1.5M & $2.0\pm 0.0$ & 200   & a       & mAP          \\
Charades~\cite{sigurdsson2016hollywood}          & 2016 & Scripted   & 81   & 6.8   & 67k  & $12.8\pm9.3$ & 157   & ML-a    & mAP          \\ \bottomrule
\end{tabular}%
} 
\caption{Action recognition results on Val (using Train) and Test (using Train+Val).}
  \resizebox{\textwidth}{!}{%
\setlength{\tabcolsep}{7pt}
\begin{tabular}{@{}clSSS SSS SSS SSS@{}}
\toprule
                                           &                                          & \multicolumn{6}{c}{Overall}                           & \multicolumn{3}{c}{Unseen Participants} & \multicolumn{3}{c}{Tail Classes} \\
                                                                                      \cmidrule(r){3-8}                                       \cmidrule(lr){9-11}                   \cmidrule(l){12-14}
                                           &                                          & \multicolumn{3}{c}{Top-1 Accuracy (\%)} & \multicolumn{3}{c}{Top-5 Accuracy (\%)} & \multicolumn{3}{c}{Top-1 Accuracy (\%)}           & \multicolumn{3}{c}{Top-1 Accuracy (\%)}    \\
                                                                                      \cmidrule(r){3-5}           \cmidrule(lr){6-8}          \cmidrule(lr){9-11}                   \cmidrule(l){12-14}
Split                                      & Baseline                              & \multicolumn{1}{c}{Verb} & \multicolumn{1}{c}{Noun} & \multicolumn{1}{c}{Act.} & \multicolumn{1}{c}{Verb} & \multicolumn{1}{c}{Noun} & \multicolumn{1}{c}{Act.}& \multicolumn{1}{c}{Verb} & \multicolumn{1}{c}{Noun} & \multicolumn{1}{c}{Act.} & \multicolumn{1}{c}{Verb} & \multicolumn{1}{c}{Noun} & \multicolumn{1}{c}{Act.} \\ \midrule
\multirow{6}{*}{\rotatebox{90}{\textbf{Val}}}
                                           & Chance                                    & 10.42 & 01.70 & 00.51   & 38.43 & 08.14 & 02.54   & 10.59 & 01.88 & 00.57   & 01.13 & 00.31 & 00.10 \\
                                           & TSN~\cite{wang2016tsn}                    & 60.18 & 46.03 & 33.19 & 89.59 & 72.90 & 55.13 & 47.42 & 38.03 & 23.47 & 30.45 & 19.37 & 13.88 \\
                                           & TRN~\cite{zhou2017trn}                    & 65.88 & 45.43 & 35.34 & 90.42 & 71.88 & 56.74 & 55.96 & 37.75 & 27.70 & 34.66 & 17.58 & 14.07 \\
                                           & TBN~\cite{kazakos2019epic}                & 66.00 & 47.23 & 36.72 & 90.46 & 73.76 & 57.66 & 59.44 & 38.22 & 29.48 & 39.09 & 24.84 & 19.13 \\
                                           & TSM~\cite{lin2019tsm}                     & 67.86 & 49.01 & 38.27 & 90.98 & 74.97 & 60.41 & 58.69 & 39.62 & 29.48 & 36.59 & 23.37 & 17.62 \\
                                           & SlowFast~\cite{feichtenhofer2019slowfast} & 65.56 & 50.02 & 38.54 & 90.00 & 75.62 & 58.60 & 56.43 & 41.50 & 29.67 & 36.19 & 23.26 & 18.81 \\ \cmidrule{1-14}
\multirow{6}{*}{\rotatebox{90}{\textbf{Test}}}
                                           & Chance                                    & 10.68 & 01.79 & 00.55   & 37.71 & 08.35 & 02.69   & 09.37 & 01.90 & 00.59   & 00.97 & 00.39 & 00.12 \\
                                           & TSN~\cite{wang2016tsn}                    & 59.03 & 46.78 & 33.57 & 87.55 & 72.10 & 53.89 & 53.11 & 42.02 & 27.37 & 26.23 & 14.73 & 11.43 \\
                                           & TRN~\cite{zhou2017trn}                    & 63.28 & 46.16 & 35.28 & 88.33 & 72.32 & 55.26 & 57.54 & 41.36 & 29.68 & 28.17 & 13.98 & 12.18 \\
                                           & TBN~\cite{kazakos2019epic}                & 62.72 & 47.59 & 35.48 & 88.77 & 73.08 & 56.34 & 56.69 & 43.65 & 29.27 & 30.97 & 19.52 & 14.10 \\
                                           & TSM~\cite{lin2019tsm}                     & 65.32 & 47.80 & 37.39 & 89.16 & 73.95 & 57.89 & 59.68 & 42.51 & 30.61 & 30.03 & 16.96 & 13.45 \\
                                           & SlowFast~\cite{feichtenhofer2019slowfast} & 63.79 & 48.55 & 36.81 & 88.84 & 74.49 & 56.39 & 57.66 & 42.55 & 29.27 & 29.65 & 17.11 & 13.45 \\ \bottomrule
\end{tabular}%

}
\label{tab:ar_results}
\end{table*}

\chParagraph{Related Datasets}
Several datasets have been collected to focus on action recognition, from~\cite{soomro2012ucf101,kuehne2011hmdb} to recent large-scale ones~\cite{gu2018ava,Kay2017,monfort2020moments,Goyal2017,zhao2019hacs,sigurdsson2016hollywood}, all offering a challenge with a held-out test set.
In Table~\ref{tab:ar_comparison}, we compare \newDataset{} to these non-egocentric datasets across a range of facets.
Ours is the only dataset of unscripted activities, of comparable size to those collected from scripted or curated (YouTube) videos.

\chParagraph{Evaluation Metrics}
We report Top-1/5 Accuracy on Val and Test sets.

\begin{figure*}[t]
    \centering
    \includegraphics[width=\textwidth]{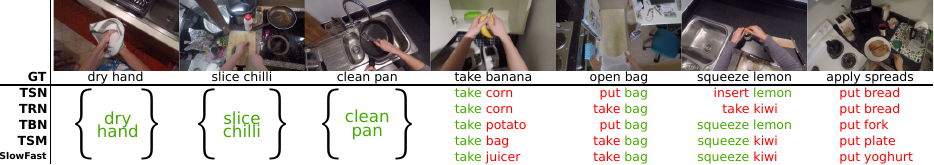}
    \caption{Qualitative action recognition results for various baselines}
    \label{fig:action_recognition}
\end{figure*}

\chParagraph{Baselines and Results}
In Table~\ref{tab:ar_results}, we report results of five state-of-the-art recognition models~\cite{wang2016tsn,zhou2017trn,kazakos2019epic,lin2019tsm,feichtenhofer2019slowfast} in addition to a random chance baseline.
We use the Train set to report on Val, optimising hyper-parameters. We then fix these, and train on both the Train and Val sets in order to report on the Test set.
Fig.~\ref{fig:action_recognition} shows success and failure examples, using examples from the Val set.

\begin{table*}[t]
\centering
\caption{Characteristics of popular datasets related to our challenges: weakly-supervised action recognition (WS), anticipation (Ant.) and detection (Det.).} 
\label{tab:weak_sup_datasets}
\resizebox{\textwidth}{!}{%
\setlength{\tabcolsep}{3pt}
\begin{tabular}{@{}lcccrrrrrrrrr@{}}
\toprule
\textbf{Dataset} & 
\textbf{WS} & 
\textbf{Ant.} & 
\textbf{Det.} & 
\textbf{Classes} & 
\textbf{Instances} & 
\textbf{Hours} & 
\textbf{\begin{tabular}[c]{@{}c@{}}Avg. Action\\Length\end{tabular}} &
\textbf{\begin{tabular}[c]{@{}c@{}}Avg. Instances\\per Video\end{tabular}} &
\textbf{\begin{tabular}[c]{@{}c@{}}Avg. Classes\\per Video\end{tabular}} &
\textbf{\begin{tabular}[c]{@{}c@{}}Labelled\\Frames\end{tabular}} &
\textbf{\begin{tabular}[c]{@{}c@{}}Overlapping\\Segments\end{tabular}} \\ \midrule
TV Human Interactions~\cite{patron2010high} & \xmark &\cmark &\xmark & 4 & 300 & 0.3  & 3.9s & 
1.0 & 1.0 & 100.0\% & 0.0\% \\
TV Series~\cite{de2016online} & \xmark &\cmark &\xmark  & 30 & 6,231 & 16.0 & 
1.9s & 
230.7 & 23.5 & N/A  & 21.9\%\\ 
THUMOS 14~\cite{THUMOS14} & \cmark & \cmark & \cmark & 101 & 6,310 & 30.0 & 4.2s & 
15.4 & 1.1 & 29.5\% & 8.3\% \\
Multi-THUMOS~\cite{yeung2015every} & \cmark & \xmark & \cmark & 65 & 38,690 & 30.0 & 3.5s & 
102.1 & 10.5 & 78.6\% & 67.1\% \\
ActivityNet 1.3~\cite{caba2015activitynet}&
 \cmark & \xmark & \cmark & 200 & 23,064 & 648.0 & 49.2s & 
1.5 & 1.0 & 65.3\% & 1.0\% \\ 
Charades~\cite{sigurdsson2016hollywood} & \cmark & \xmark & \cmark & 157 & 66,500 & 81.1 & 12.8s 
& 6.9 & 6.8 & 89.7\% & 79\% \\
UCF 101-24~\cite{THUMOS15} & \cmark & \xmark & \cmark & 24 & 4,569 & 6.2 & 5.1s 
& 1.4 & 1.0 & 82.5\% & 18.8\% \\
DALY~\cite{weinzaepfel2016human} & \cmark & \xmark & \cmark & 10 & 3,907 & 31.9 & 7.7s 
& 7.6 & 1.1 & 30.5\% & 8.4\% \\ 
 Hollywood2~\cite{marszalek2009actions} & \cmark & \xmark & \xmark & 16 & 2,376 & 21.0 & 4.3s 
& 2.5 & 1.9 & 37.0\% & 0.0\% \\
Breakfast~\cite{Kuehne2014} (cam01) & \cmark & \cmark & \cmark & 8 & 9,941 & 77.0 & 3.0s 
& 38.7 & 18.5 & 96.6\% & 0.4\% \\
50 Salads~\cite{Stein2013} & \cmark & \cmark & \cmark & 17 & 899 & 4.5 & 36.8s 
& 18.0 & 16.3 & 85.9\% & 3.6\% \\
MPII~\cite{Rohrbach2012} & \cmark & \cmark & \cmark & 65 & 5,609 & 8.3 
& 11.1m & 127.5 & 24.9 & 97.3\% & 0.1\%  \\
EGTEA Gaze+~\cite{EGTEA} & \xmark & \cmark & \cmark & 106 & 15,484 & 28.0 & 3.7s 
& 180.0 & 82.8 & 57.9\% & 6.1\% \\\midrule
\newDataset & \cmark & \cmark & \cmark & 4,053  & 89,977 & 100.0 & 3.1s 
& 128.5 & 53.2  & 71.6\% & 28.1\%
\\ \bottomrule
\end{tabular}%
}
\end{table*}

\begin{table*}
\caption{Weakly-supervised action recognition results.}
\label{tab:single_ts_results}
\setlength{\tabcolsep}{7pt}
\resizebox{\linewidth}{!}{%
\begin{tabular}{@{}clcccccccccccc@{}}
\toprule
 &  & \multicolumn{6}{c}{Overall} & \multicolumn{3}{c}{Unseen Participants} & \multicolumn{3}{c}{Tail Classes} \\ \cmidrule(r){3-8} \cmidrule(lr){9-11} \cmidrule(l){12-14} 
 &  & \multicolumn{3}{c}{Top-1} & \multicolumn{3}{c}{Top-5} & \multicolumn{3}{c}{Top-1} & \multicolumn{3}{c}{Top-1} \\ \cmidrule(r){3-5} \cmidrule(lr){6-8} \cmidrule(lr){9-11} \cmidrule(l){12-14}  
Split & Baseline & Verb & Noun & Act. & Verb & Noun & Act. & Verb & Noun & Act. & Verb & Noun & Act. \\ \midrule
\multirow{2}{*}{\rotatebox{90}{\textbf{Val}}} & Fixed segment \hspace{6em} & 44.86 & 37.97 & 20.30 & 84.62 & 65.41 & 39.35 & 37.37 & 29.20 & 14.36 & 25.85 & 18.89 & 10.50 \\
 &\cite{Moltisanti_2019_CVPR} & 47.18 & 38.23 & 22.24 & 85.66 & 66.20 & 40.87 & 40.94 & 30.33 & 17.56 & 27.10 & 19.31 & 10.86 \\ \midrule
\multirow{2}{*}{\rotatebox{90}{\textbf{Test}}} & Fixed segment & 43.93 & 38.01 & 20.38 & 82.54 & 65.85 & 39.25 & 40.70 & 34.79 & 18.17 & 21.26 & 13.57 & 07.18  \\
 &\cite{Moltisanti_2019_CVPR} & 46.59 &  37.33 & 21.79 & 82.97 & 65.78 & 40.83 & 42.80 & 32.29 & 18.37 & 21.81 & 14.28 & 08.23 \\ \bottomrule 
\end{tabular}%
}

\end{table*}

\subsection{Weakly-supervised Action Recognition}
\label{sec:action_weakly_recognition_challenge}

\chParagraph{Definition}
\edits{As in Sec~\ref{sec:action_recognition_challenge}, the goal is to recognise the action, \textit{i.e.} predict $(\hat{v}, \hat{n}, \hat{a})$, in a 
\textit{trimmed} action segment during \textit{testing}.
Distinctly, we use single timestamps instead of temporal boundaries during \textit{training}.}
Let $\mathcal{A}=(A_i)_{i=1}^N$ be the action instances contained in \edits{an untrimmed training} video, each ${A_i=(t, v, n, a)}$ is labelled with only one timestamp $t$ roughly located around the action, along with verb/noun classes.
We utilise the narration timestamps from our collection pipeline as $t$.

\chParagraph{Related Datasets and Types of Supervision}
Previous weakly-supervised approaches utilised video-level or transcript supervision, where the set~\cite{wang2017untrimmednets,singh2017hide,nguyen2018weakly,liu2019completeness,nguyen2019weakly,narayan20193c}
 or sequence~\cite{bojanowski2014weakly,huang2016connectionist,ding2018weakly,richard2018neuralnetwork,chang2019d3tw,li2019weakly} of actions in the video are used in training, without temporal bounds.
Table~\ref{tab:weak_sup_datasets} compares \newDataset{} to datasets trained with weak-supervision. 
When considering the number of classes (and instances) per video, \newDataset{} offers a significant challenge.
For example, ActivityNet~\cite{caba2015activitynet} videos contain 1 class and 1.5 action instances on average, whereas in \newDataset{}, videos contain 53.2 classes and 128.5 instances.
Video-level supervision is only sufficient for datasets with a few classes per video~\cite{caba2015activitynet,THUMOS14}, while transcript supervision~\cite{marszalek2009actions,Kuehne2014} expects no overlap between actions.
Both types of weak supervision are insufficient in our case.

Alternatively, single-timestamp supervision is gaining popularity due to the scalability and performance balance~\cite{Moltisanti_2019_CVPR,bearman2016s,mettes2016spot,cheron2018flexible}.
We follow this trend as it fits naturally with our narration timestamps collected using \annStyle.

\chParagraph{Evaluation Metrics} We follow the same metrics as in Section~\ref{sec:action_recognition_challenge}. 

\chParagraph{Baselines and Results} 
We consider two baselines. 
The first, ``Fixed segment'', uses a segment of fixed length centred on the timestamp. 
The second is our previous work~\cite{Moltisanti_2019_CVPR}, where sampling distributions\edits{, to select training frames from the untrimmed videos,} are initialised from single timestamps, and refined based on the classifier's response\footnote{\edits{The distributions are modelled with a plateau function, initialised with a fixed width and slope, and centred around the annotated timestamp.
These are refined from the classification scores iteratively. More details in~\cite{Moltisanti_2019_CVPR}}}.
Both are trained end-to-end using a TSN backbone~\cite{wang2016tsn} and results can be seen in Table~\ref{tab:single_ts_results}. \cite{Moltisanti_2019_CVPR}~improves the fixed segment baseline by 1-3\% top-1 accuracy across Val and Test. The~fully supervised upper bound is TSN, reported in Table~\ref{tab:ar_results}. Comparatively, weak supervision performs 11\% worse than strong supervision on top-1 action accuracy in Val and Test. 
Using roughly aligned single timestamps is
challenging when actions are short and overlapping. 
\newDataset, with its dense actions, provides an interesting benchmark to develop new models for weak-supervision.

\subsection{Action Detection}
\label{sec:action_detection_challenge}

\chParagraph{Definition}
All other challenges in Section~\ref{sec:challenges} consider a trimmed segment $(t_s, t_e)$ from the test video as input. This assumption is limiting, as labelled start/end times of actions are unlikely to be present for new test videos. In this challenge, we aim to detect and recognise all action instances within an untrimmed video, as in~\cite{caba2015activitynet}. 
Given a video, we predict the set of all action instances $\hat{\mathcal{A}}=\{\hat A_i\}_{i=1}^M$, where $\hat A_i = (\hat t_s, \hat t_e, \hat v, \hat n, \hat{a}$) is an action detection tuple including the predicted start and end times $(\hat t_s, \hat t_e)$ and the predicted classes $(\hat v, \hat n, \hat{a})$. During training, we use the set of ground-truth action annotations~$\mathcal{A}$. Note that the ground-truth $\mathcal{A}$ and predicted $\hat{\mathcal{A}}$ sets can be of different sizes.
This definition is closely related to temporal segmentation~\cite{lea2017temporal}, but segmentation assumes non-overlapping segments and is thus unsuitable for \newDataset{}.

\chParagraph{Related Datasets} 
Table~\ref{tab:weak_sup_datasets} compares \newDataset\ to popular datasets for temporal action detection and segmentation. 
\newDataset\ presents the largest challenge, when considering the combined metrics of: average video length, average instances per video and overlapping instances. 
Compared to datasets with overlapping segments, it has a larger number of instances per video and  is also longer (in hours) than all datasets with higher average instances per video.

\chParagraph{Evaluation Metrics} 
In line with~\cite{caba2015activitynet}, we use mean Average Precision (mAP) by computing the average of the AP values for each class. 
A predicted segment matches a ground truth segment if their Intersection over Union (IoU) is greater than or equal to thresholds ranging from $0.1$ to $0.5$.

\begin{figure*}[t]
    \centering
    \includegraphics[width=\linewidth]{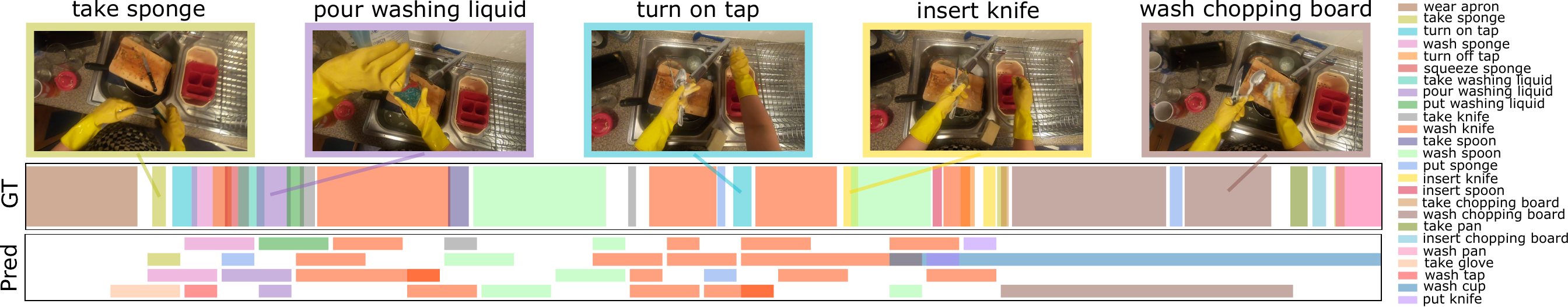}
    \caption{Qualitative results of action detection. \edits{Predictions with confidence $> 0.5$ are shown with colour-coded class labels (see legend). Since the baseline predicts overlapping segments, the predictions are displayed over four rows for ease of viewing.}}
    \label{fig:tad_qualitative}
\end{figure*}

\chParagraph{Baselines and Results}
We consider a two-stage baseline. Action proposals are first obtained using Boundary Matching \edits{Networks} (BMN)~\cite{lin2019bmn}, which are then classified using SlowFast~\cite{feichtenhofer2019slowfast} (model trained as in Section~\ref{sec:action_recognition_challenge}).
Results in Table~\ref{tab:tad_results} highlight that action detection is particularly challenging on this dataset, especially with respect to higher IoU thresholds. The qualitative example in Fig.~\ref{fig:tad_qualitative} shows that our videos in \newDataset{} contain actions of varying lengths, which adds further challenges.

\begin{table}[t!]
	\caption{Temporal action detection results in mAP (\%).}
	\label{tab:tad_results}
	\resizebox{\linewidth}{!}{%
		\setlength{\tabcolsep}{4pt}
		\begin{tabular}{@{}cllcccccc@{}}
			\toprule
			&&& \multicolumn{6}{c}{Mean Average Precision (mAP)} \\
			\cmidrule{4-9}
			Split & Baseline & Task & 
			\multicolumn{1}{c}{@0.1} & \multicolumn{1}{c}{@0.2} & \multicolumn{1}{c}{@0.3} & \multicolumn{1}{c}{@0.4} & \multicolumn{1}{c}{@0.5} & \multicolumn{1}{c}{Avg.} \\ \midrule
			
			\multirow{3}{*}{\rotatebox{90}{\textbf{Val}}}
			&
			\multirow{3}{*}{\shortstack{BMN~\cite{lin2019bmn} +\\ SlowFast~\cite{feichtenhofer2019slowfast}}}
			&Verb & 10.83 & 09.84 & 08.43 & 07.11 & 05.58 & 08.36\\
			&&Noun & 10.31 & 08.33 & 06.17 & 04.47 & 03.35 & 06.53\\
            &&Act. & 06.95 & 06.10 & 05.22 & 04.36 & 03.43 & 05.21\\
			\midrule
\multirow{3}{*}{\rotatebox{90}{\textbf{Test}}}&
\multirow{3}{*}{\shortstack{BMN~\cite{lin2019bmn} +\\ SlowFast~\cite{feichtenhofer2019slowfast}}}
&Verb & 11.10 & 09.40 & 07.44 & 05.69 & 04.09 & 07.54\\
&&Noun & 11.99 & 08.49 & 06.04 & 04.10 & 02.80 & 06.68\\
&&Act. & 06.40 & 05.37 & 04.41 & 03.36 & 02.47 & 04.40\\

			\bottomrule
		\end{tabular}%
		
	}

\end{table}

\begin{table*}[t!]
\caption{Action anticipation results reported in class-mean top-5 recall (\%).}
  \resizebox{\textwidth}{!}{%
\setlength{\tabcolsep}{8pt}
\begin{tabular}{@{}clccccccccc@{}}
\toprule
                                           &                        & \multicolumn{3}{c}{Overall} & \multicolumn{3}{c}{Unseen Participants} & 
                                           \multicolumn{3}{c}{Tail Classes}    \\
                                                                    \cmidrule(r){3-5}           \cmidrule(lr){6-8}          \cmidrule(l){9-11} 
Split                                      & Baseline                  & \multicolumn{1}{c}{Verb} & \multicolumn{1}{c}{Noun} & \multicolumn{1}{c}{Act.} & \multicolumn{1}{c}{Verb} & \multicolumn{1}{c}{Noun} & \multicolumn{1}{c}{Act.}& 
\multicolumn{1}{c}{Verb} & \multicolumn{1}{c}{Noun} & \multicolumn{1}{c}{Act.} \\ \midrule
\multirow{2}{*}{\rotatebox{90}{\textbf{Val}}}
                                           
&chance&06.39 & 02.00 & 00.20 & 14.35 & 02.88 & 00.51 & 01.64 & 00.24 & 00.05
 \\ 

&RU-LSTM~\cite{furnari2019would} \hspace*{60pt}
&27.76 & 30.76 & 14.04 & 28.78 & 27.22 & 14.15 & 19.77 & 22.02 & 11.14\\

                                           \cmidrule{1-11}
\multirow{2}{*}{\rotatebox{90}{\textbf{Test}}}
&chance &06.17 & 02.28 & 00.14 & 08.14 & 03.28 & 00.31 & 01.87 & 00.66 & 00.03

 \\
                  &RU-LSTM~\cite{furnari2019would}&
                  25.25 & 26.69 & 11.19 & 19.36 & 26.87 & 09.65 & 17.56 & 15.97 & 07.92\\

                                           \bottomrule
\end{tabular}%
}
\label{tab:aa_results}
\end{table*}

\subsection{Action Anticipation}
\label{sec:action_anticipation_challenge}
\chParagraph{Definition}  
We aim to predict ($\hat{v},\hat{n},\hat{a}$) as the verb/noun/action classes of the action, by observing a video segment \edits{of arbitrary duration $\tau_o$ seconds (observation time) ending $\tau_a$ seconds (anticipation time) before} the action's start, $t_s$\edits{. We set $\tau_a=1$.}
\edits{We expect models addressing this task to reason on observed sequences of actions, the current state of the world (e.g., what objects are visible) and the possible goal of the camera wearer.}

\chParagraph{Related Datasets}
Table~\ref{tab:weak_sup_datasets} also compares \newDataset{} with other datasets used for action anticipation~\cite{Rohrbach2012,patron2010high,de2016online,THUMOS14,Kuehne2014,Stein2013,EGTEA}. 
Our dataset is the largest in hours and classes, and is unscripted, which is critical for meaningful anticipation models, and for in the wild testing.

\chParagraph{Evaluation Metrics}
We report results using class-mean top-5 recall~\cite{furnari2018leveraging}. The top-k criterion accounts for uncertainty in future predictions, as with previous anticipation efforts~\cite{koppula2013anticipating,lee2017desire,bhattacharyya2018bayesian}.
Class-mean allows for balancing the long-tail distribution.

\begin{figure*}[t]
    \centering
    \includegraphics[width=\textwidth]{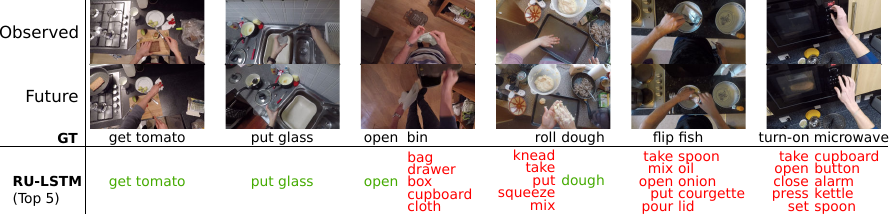}
    \caption{Qualitative action anticipation results.}
    \label{fig:action_anticipation_qualitative}
\end{figure*}

\chParagraph{Baselines and Results}
We use our prior work RU-LSTM~\cite{furnari2019would} as a baseline.
In Table~\ref{tab:aa_results}, RU-LSTM performs better for nouns compared to verbs, but shows that tail classes are particularly challenging for anticipation.
Fig.~\ref{fig:action_anticipation_qualitative} demonstrates the baseline struggles where the next active noun/verb are ambiguous.

\subsection{Unsupervised Domain Adaptation for Action Recognition}
\label{sec:domain_adaptation_challenge}
\chParagraph{Definition}
Unsupervised Domain Adaptation (UDA) utilises a labelled source domain and learns to adapt to an unlabelled target domain.
We use videos recorded in 2018 as the labelled source, and use newly collected videos as unlabelled target (i.e. without any of the accompanying annotations).
The action recognition task itself follows the definition in Section~\ref{sec:action_recognition_challenge}.
The difficulty of this challenge stems from the fact that the source and target domains come from distinct training distributions due to the collection of videos two years later \edits{. Changes in location, hardware and long-term temporal offsets are the main sources of the domain shift} (see Section~\ref{sec:statistics}).  
A~method which is able to perform this task well provides a number of practical benefits, most notably the elimination of labelling time and expense when collecting new videos, in the future.

\begin{table*}[t!]
    \begin{center}
        \caption{Comparison of domain adaptation classification datasets.  SD: Subdomains. M: Modalities. AAL: Average Action Length. *: Note that Test* refers to `Target Test'}  
    \label{tab:da_comparison}

    \resizebox{\textwidth}{!}{%
        \begin{tabular}{llrrrrrrrrrr} 
    \toprule
    & \textbf{Dataset}               & \multicolumn{2}{l}{\textbf{Train}} & \textbf{Test} & \textbf{Classes} & \textbf{SD} & \textbf{M} & \textbf{AAL} & \textbf{Year} &  \textbf{Real/Syn }\\ \midrule
    \multirow{4}{*}{\rotatebox[]{90}{\textbf{Image}}} & Office \cite{Saenko2010}         & \multicolumn{2}{r}{4110}\hspace{30pt} & N/A & 31     & 1 & 3 & N/A\hspace{30pt} & 2010   & Real\\
    & ImageCLEF~\cite{caputo2014imageclef} & \multicolumn{2}{r}{2400}\hspace{30pt} & 600 & 12 & 5 & 1 & N/A\hspace{30pt} & 2014 & Real\\
    & Office$\leftrightarrow{}$Home~\cite{Venkateswara2017}        & \multicolumn{2}{r}{15500}\hspace{30pt} & N/A & 65     & 4 & 1 & N/A\hspace{30pt} & 2017  & Real\\
    & VisDA-C \cite{Peng2017}            & \multicolumn{2}{r}{280157}\hspace{30pt}  & N/A & 12     & 3 & 1 & N/A\hspace{30pt} & 2017   & Real/Syn\\
    & DomainNet \cite{Peng2019}          & \multicolumn{2}{r}{363534}\hspace{30pt} & 37706 & 345     & 6 & 1 & N/A\hspace{30pt} & 2019   & Real/Syn\\ \midrule
     & & \textbf{Source} & \textbf{Target} & \textbf{Test*}\\
    \midrule
    \multirow{6}{*}{\rotatebox[]{90}{\textbf{Video}}} & UCF$\leftrightarrow{}$HMDB (small) \cite{xu2016dual} & 482 & 350 & 150 & 5     & 2 & 1 & $4.7 \pm 2.5$\hspace{30pt} & 2018   & Real  \\
    & UCF$\leftrightarrow{}$Olympic \cite{jamal2018deep}        & 601  & 250 & 54 & 6     & 2 & 1 & $6.6 \pm 4.5$\hspace{30pt} & 2018   & Real \\
    & UCF$\leftrightarrow{}$HMDB (full) \cite{Chen2019}    & 1438  & 840 & 360 & 12     & 2 & 1 & $4.0 \pm 5.8$\hspace{30pt} & 2019   & Real\\
    & IEMOCAP$\rightarrow{}$AFEW \cite{qi2018unified} & 6611 &  795 & N/A & 4 & 2 & 2 & N/A\hspace{30pt} & 2018  & Real \\
    & Kinetics$\leftrightarrow{}$Gameplay \cite{Chen2019}  & 43378 & 2625 & 749 & 30     & 2 & 1 & N/A\hspace{30pt} & 2019    & Real/Syn \\
    & \newDataset         & 16115 & 26115 & 5909 & 3369       & 16 & 3 & $2.8 \pm 5.2$\hspace{30pt} & 2020    & Real\\
    \bottomrule
    \end{tabular}
    }
\end{center}
\end{table*}

\chParagraph{Related Datasets}
UDA datasets have traditionally used images \cite{Saenko2010,Venkateswara2017,Peng2017,Peng2019}, with recent attempts to use video~ \cite{jamal2018deep,Chen2019,qi2018unified} adapting across public datasets (\eg UCF to Olympics). \newDataset{} is the first to propose a within-dataset domain adaptation challenge in video.
Video-based UDA raises additional challenges, such as aligning temporal information across domains~\cite{jamal2018deep}, attending to relevant transferable frames~\cite{Chen2019}, and avoiding non-informative background frames~\cite{pan2019adversarial}.  

Table~\ref{tab:da_comparison} shows 
\newDataset{} provides 
several advantages over other video-based datasets: largest number of instances, classes, subdomains, and is multi-modal~\cite{munro2020multi}.  
Additionally, it has a compound domain gaps resulting from the test of time (\ie recording data two years later).

\chParagraph{Splits}
This challenge assesses models' ability to adapt to additional footage without labels. We thus define the following splits; \textit{Source}: labelled training data from 16 participants (collected in 2018) and \textit{Target:} unlabelled footage from the same 16 participants collected in 2020. This ensures the gap in the domains is related to the capturing of the data `two years later'. We further split target videos into: \textit{Target Train} and \textit{Target Test}. The first are unlabelled videos used during domain adaptation, while the second are videos used for evaluation, as in~\cite{Peng2017}. Number of action instances per split are reported in Table~\ref{tab:da_comparison}. 

\begin{figure}[t!]
\centering
   \includegraphics[width=1\linewidth]{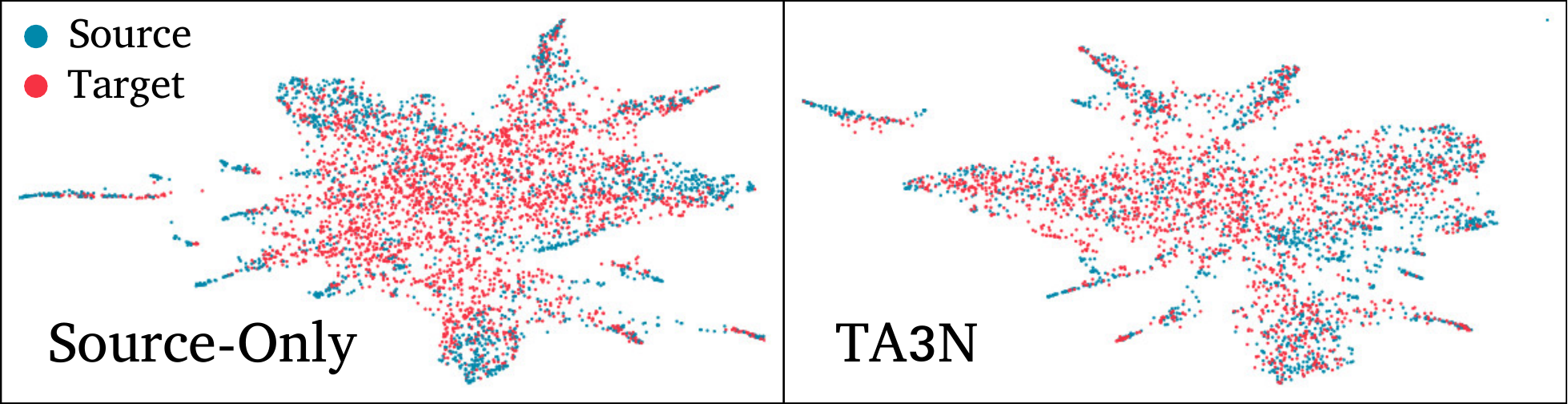}
\caption{UMAP~\cite{mcinnes2018umap} of feature spaces shows better alignment through UDA baseline.}%
\label{fig:UD_visualisation}%
\end{figure}

\chParagraph{Evaluation}
We use the same evaluation metrics as Section~\ref{sec:action_recognition_challenge} on Target Test.

\chParagraph{Baselines and Results}
We present lower and upper bounds: ``Source-Only'', where labelled source data is used for training and no adaptation to target data is attempted, and two upper bounds: ``Target-Only'', where \textit{labelled} target data is used and ``Source+Target'' where all training data is used with associated labels. Neither of these are UDA methods, but offer an insight into the domain gap.

Table~\ref{tab:DA_results} reports the results for the baselines. These use extracted features from TBN~\cite{kazakos2019epic} trained on source.
We use the code of Temporal Attentive Alignment (TA3N) \cite{Chen2019}, modified to consider multi-modal features~(RGB, Flow and Audio), to report results. 
These show significant performance improvement when using multi-modal data compared to \edits{single modality models of RGB, Flow and Audio}.
The domain gap is evident when comparing the lower and upper bounds. 
TA3N is able to partially decrease this gap, providing a $2.5\%$ improvement in verb accuracy and 2.4\% in nouns when using multiple modalities. 
\edits{Recent work~\cite{planamente2021cross} showed that RGB and Audio exhibit different levels of robustness to the domain gap in \newDataset. The best performing submissions for this challenge in 2021 exploited multi-modalities for domain adaptation~\cite{yang2021epic,plizzari2021polito}.} 
Fig.~\ref{fig:UD_visualisation} visualises the multi-modal feature space showing limited overlap between source and target. TA3N aligns the features demonstrating the capability of UDA. 

\begin{table}
        \caption{Unsupervised domain adaptation results with lower (source-only) and \edits{the upper bounds of target-only and source+target}. *modified to consider multi-modal features}
    \label{tab:DA_results}
    \begin{center}
        \resizebox{\linewidth}{!}{%
        \setlength{\tabcolsep}{2pt}
    \begin{tabular}{llrrrrrr}
    \toprule
        & & \multicolumn{3}{c}{Top-1 Acc. (\%)} & \multicolumn{3}{c}{Top-5 Acc. (\%)} \\
        \cmidrule(lr){3-5}
        \cmidrule(lr){6-8}
        Modality & Baseline & Verb & Noun & Act. & Verb & Noun & Act. \\
        \midrule
         \multirow{3}{*}{RGB} & Source-Only & 32.8 & 21.2 & 10.7 & 72.6 & 43.9 & 22.6\\
         & TA3N~\cite{Chen2019} & 32.1 & 21.6 & 11.1 & 71.7 & 44.1 & 22.5\\
         &\cellcolor{lightgray} Target-Only & \cellcolor{lightgray} 39.7 & \cellcolor{lightgray} 32.3 & \cellcolor{lightgray} 18.3 & \cellcolor{lightgray} 80.8 & \cellcolor{lightgray} 56.2 & \cellcolor{lightgray} 34.0 \\
          &\cellcolor{lightgray} Source+Target & \cellcolor{lightgray} 41.1 & \cellcolor{lightgray} 33.0 & \cellcolor{lightgray} 18.8 & \cellcolor{lightgray} 80.4 & \cellcolor{lightgray} 58.5 & \cellcolor{lightgray} 35.2 \\
          \midrule
         \multirow{3}{*}{Flow} & Source-Only & 42.8 & 19.2 & 12.7 & 74.5 & 38.5 & 23.8\\
         & TA3N~\cite{Chen2019} & 43.2 & 20.1 & 12.8 & 74.5 & 41.2 & 25.0\\
         &\cellcolor{lightgray} Target-Only & \cellcolor{lightgray} 53.8 & \cellcolor{lightgray} 26.7 & \cellcolor{lightgray} 20.2 & \cellcolor{lightgray} 84.2 & \cellcolor{lightgray} 49.1 & \cellcolor{lightgray} 34.5 \\
          &\cellcolor{lightgray} Source+Target & \cellcolor{lightgray} 52.4 & \cellcolor{lightgray} 27.3 & \cellcolor{lightgray} 19.8 & \cellcolor{lightgray} 82.4 & \cellcolor{lightgray} 50.3 & \cellcolor{lightgray} 35.4 \\
          \midrule
         \multirow{3}{*}{Audio} & Source-Only & 31.4 & 12.8 & 8.5 & 64.8 & 28.4 & 16.0\\
         & TA3N~\cite{Chen2019} & 32.0 & 13.3 & 8.9 & 66.0 & 29.1 & 16.5\\
         &\cellcolor{lightgray} Target-Only & \cellcolor{lightgray} 41.7 & \cellcolor{lightgray} 19.1 & \cellcolor{lightgray} 13.4 & \cellcolor{lightgray} 77.2 & \cellcolor{lightgray} 39.6 & \cellcolor{lightgray} 23.6 \\
          &\cellcolor{lightgray} Source+Target & \cellcolor{lightgray} 41.8 & \cellcolor{lightgray} 19.8 & \cellcolor{lightgray} 13.8 & \cellcolor{lightgray} 77.1 & \cellcolor{lightgray} 40.6 & \cellcolor{lightgray} 24.3 \\
          \midrule
         \multirow{3}{*}{\shortstack{RGB+Flow\\+Audio}}& Source-Only & 44.4 & 25.3 & 16.8 & 69.7 & 48.4 & 29.1 \\
         & TA3N*~\cite{Chen2019} & 46.9 & 27.7 & 19.0 & 72.7 & 50.7 & 30.5\\
         & \cellcolor{lightgray} Target-Only & \cellcolor{lightgray} 59.1 & \cellcolor{lightgray} 40.3 & \cellcolor{lightgray} 30.4 & \cellcolor{lightgray} 85.0 & \cellcolor{lightgray} 65.0 & \cellcolor{lightgray} 47.8\\
         & \cellcolor{lightgray} Source+Target & \cellcolor{lightgray} 59.4 & \cellcolor{lightgray} 41.9 & \cellcolor{lightgray} 31.3 & \cellcolor{lightgray} 85.3 & \cellcolor{lightgray} 66.6 & \cellcolor{lightgray} 49.2\\
    \bottomrule
    \end{tabular}}
    \end{center}
    
\end{table}

\subsection{Multi-Instance Action Retrieval}
\label{sec:action_retrieval_challenge}

\chParagraph{Definition}
\label{subsec:act_retr_definition}
Given a query action segment, the aim of video-to-text retrieval is to rank captions in a gallery set, $C$, such that those with a higher rank are more \emph{semantically relevant} to the action in the video. 
Conversely, text-to-video retrieval uses a query caption $c_i \in C$ to rank videos. 
Different from other challenges in Section~\ref{sec:challenges}, we here use the English-translated free-form captions from the narrations~(Fig.~\ref{fig:main_annotation}b). 

\chParagraph{Splits}We use the Train split from Table~\ref{tab:dataset_split_stats}. As access to the captions are required for both video-to-text and text-to-video retrieval, the Val set is used for evaluating this challenge to allow the held-out Test set for all other challenges to remain intact. We consider all the videos in Val, and all unique captions, removing repeats.

\begin{table*}[t!]
\centering
\caption{Multi-Instance retrieval datasets.}
\label{tab:action_retrieval_datasets}
\resizebox{\textwidth}{!}{%
\begin{tabular}{@{}lrrrrrrrr@{}}
\toprule
\textbf{Dataset} & \textbf{Segments} & \textbf{Cap.} & \textbf{Cap. Length } & \textbf{Cap./Seg.} & \textbf{Action Length} & \textbf{Rel. Caps.} & \textbf{Vid. Source} & \textbf{Cap. Source}\\ \midrule
MSR-VTT~\cite{xu2016msr} & 10,000 & 200,000 & 9.3 & 20.0 & 15.0s & 20.0 & YouTube & Collected \\
MSVD~\cite{chen2011collecting} & 2,089 & 122,665 & 7.1 & 40.9 & 9.8s & 40.9 & YouTube & Collected \\
LSMDC~\cite{rohrbach2015dataset} & 69,000 & 68,000 & 9.6 & 1.0 & 3.9s & 1.0 & Film & Script/AD \\
YouCook2~\cite{zhou2017procnets} & 13,829 & 13,829 & 8.8 & 1.0 & 19.6s & 1.0 & YouTube & Subtitle \\ \midrule
\newDataset & 76,885& 19,803 & 3.0 & 1.0 & 3.1s & 22.4 & Unscripted & Narration \\ \bottomrule
\end{tabular}%
}

\end{table*}

\begin{figure*}[t]
    \centering
    \includegraphics[width=\textwidth]{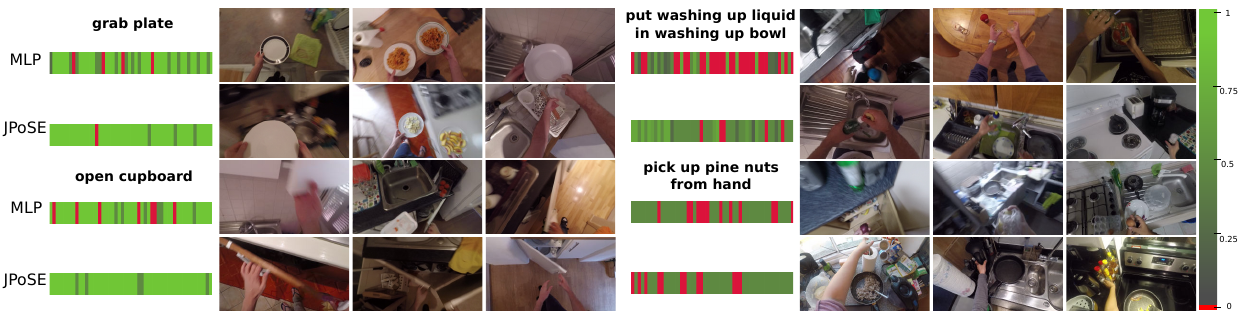}
    \caption{Qualitative results for text-to-video action retrieval. Top 3 retrieved videos and the semantic relevancy $\mathcal{R}$ of the top 50 retrievals (red: irrelevant, green: relevant).}
    \label{fig:action_retrieval}
\end{figure*}

\begin{table}        \caption{Multi-Instance retrieval results.}
    \label{tab:retr_chall_results}
    \resizebox{\linewidth}{!}{%
     \setlength{\tabcolsep}{2pt}
    \begin{tabular}{lccccccc}
    \toprule & \multicolumn{3}{c}{mAP}&& \multicolumn{3}{c}{nDCG} \\ \cmidrule{2-4} \cmidrule{6-8}
     Baseline & {vid$\rightarrow$txt}  & {txt$\rightarrow$vid}  & {Avg.} && {vid$\rightarrow$txt} & {txt$\rightarrow$vid} & {Avg.} \\ \midrule
    Chance                                    & 5.7 & 5.6 & 5.7 && 10.8 &10.9 &10.9 \\ \midrule
    MLP                      &43.0 & 34.0&38.5 && 50.1 &46.9 &48.5 \\
    JPoSE~\cite{wray2019fine}  &49.9 &38.1 &44.0 && 55.5 &51.6 &53.5 \\ \bottomrule
    \end{tabular}}%
\end{table}

\chParagraph{Related Datasets}
In datasets that are commonly used for retrieval~\cite{xu2016msr,rohrbach2015dataset,zhou2017procnets,chen2011collecting}, 
captions are considered relevant if they were collected for the same video, and irrelevant otherwise. This common approach ignores the semantic overlap between captions of different videos that contain identical or similar actions. 
These datasets thus assume videos to be distinct from one another.
In instructional video datasets~\cite{zhou2017procnets,miech2019howto100m}, the corresponding YouTube subtitle is only considered relevant, again ignoring semantic overlap or similarities to other actions. \edits{Note that the large-scale HowTo100M~[117] dataset has only been used for pre-training, due to being webly supervised and thus noisy. The dataset does not include a val/test set.}

In this challenge, we use the class knowledge from Section~\ref{sec:statistics} to define caption relevancy. 
This allows us to consider captions ``put glass'' and ``place cup'' as semantically relevant---an opportunity not available in other retrieval datasets.

\chParagraph{Evaluation Metrics}
To evaluate this challenge, relevancy of a retrieved caption (or video) to the query item needs to be assessed.
We consider the case where a query video contains the action of \emph{someone cutting a pizza using a cutter}.
We want captions: a) ``cutting a pizza using a cutter'', b) ``cutting a pizza slice'', c)~``slicing a pizza'' to all be \textit{more} relevant than d) ``cutting a pizza using a knife'' which in turn is \textit{more} relevant than both e) ``cutting a vegetable'' or f) ``picking up a pizza slice''.
Critically, g) ``opening a fridge'' should be considered irrelevant.

Mean Average Precision (mAP) has been used in other retrieval works~\cite{wray2019fine,rasiwasia2014cluster,kang2015learning,cao2016correlation}, yet it only considers relevance between items to be binary.
Because of this, (a--c) would be considered (equally) relevant captions.
However, we would also like to consider non-binary relevance where (d) is more relevant than~(e) which in turn is more relevant than (g).
We thus also report results using normalised Discounted Cumulative Gain (nDCG)~\cite{jarvelin2002cumulated}. This metric allows for non-binary relevance between captions.
We define the relevance, $\mathcal{R}$, as the mean IoU of the verb and noun classes, giving a value between 0 and 1, where 0 is irrelevant (no overlap in verb/noun classes) and 1 is extremely relevant.
From the example above, 1 = $\mathcal{R}$(a,a) $\ge$ $\mathcal{R}$(a,b) = $\mathcal{R}$(a,c) $\ge$ $\mathcal{R}$ (a,d) $\ge$ $\mathcal{R}$ (a,e) = $\mathcal{R}$ (a,f) $\ge$ $\mathcal{R}$ (a,g) = 0.
We then use $\mathcal{R}$ to calculate nDCG as in~\cite{jarvelin2002cumulated} (see appendix~\ref{sec:appendix-retrieval} for full definition).

\chParagraph{Baselines and Results}
As in Section~\ref{sec:domain_adaptation_challenge}, we use TBN~\cite{kazakos2019epic} features trained on the Train split.
Table~\ref{tab:retr_chall_results} provides results for two baselines and the chance lower bound.
Multi-Layer Perceptron (MLP) uses a 2-layer perceptron to project both modalities into a shared action space with a triplet loss. Our previous work JPoSE~\cite{wray2019fine} disentangles captions into verb, noun and action spaces learned with a triplet loss.
JPoSE sees a significant boost in performance over MLP.
Fig.~\ref{fig:action_retrieval} shows qualitative retrieval results on four examples using both MLP and JPoSE for text-to-video retrieval. JPoSE is able to retrieve more correct videos than MLP, but both methods still struggle on longer captions.
Importantly, this dataset offers the first opportunity for  action retrieval that considers semantic similarity.

\section{Conclusion and Future Work}
We presented our large-scale egocentric dataset \newDataset{}, through an annotation pipeline that is scalable and \edits{is} of higher quality than previous approaches. We defined six challenges, providing leaderboard baselines. Dataset and leaderboards are available at \textcolor{blue}{\underline{\url{http://epic-kitchens.github.io}}}.

These 6 challenges have been chosen to facilitate progress in open topics within video understanding. They also highlight interesting parts of our collection and annotation pipeline. For example, retrieval uses our free-form captions, while unsupervised domain adaptation for action recognition builds on collecting footage two years later. Our dense annotations of overlapping actions make detection in long untrimmed videos particularly challenging. While this paper addresses each challenge independently, successful methods that address one challenge~(\eg~detection) are likely to prove advantageous for better performance in another~(\eg~anticipation). Combining all challenges with unsupervised domain adaptation would enable future deployment in new environments without additional labels.

In publishing this manuscript we hope that people can not only utilise this large-scale dataset in their ongoing research, but also build on our novel pipeline in collecting our dataset. The proposed \annStyle{} narrator, publicly available, as well as our visually-supported transcription interfaces can prove \edits{advantageous} for other large-scale collection efforts.

\noindent \textbf{Data Release Statement:}
Dataset sequences, extracted frames and optical flow are available under Non-Commercial Government Licence for public sector information at the University of Bristol data repository: \linebreak
\resizebox{\linewidth}{!}{\textcolor{blue}{\underline{\url{http://dx.doi.org/10.5523/bris.2g1n6qdydwa9u22shpxqzp0t8m}}}}

\noindent Annotations, models, evaluation scripts, challenge leaderboards and updates are available at: \linebreak 
\textcolor{blue}{\underline{\url{http://epic-kitchens.github.io}}}

\bibliographystyle{splncs}
\bibliography{library}

\section*{Appendices}
\appendix

\section{Video Demonstration}
We provide a video demonstration of our annotation pipeline and six challenges.
Our video utilises a single sequence, showcasing the annotation pipeline first, as the sequence progresses.
We demonstrate the `pause-and-talk' narrator, transcription and translation steps, then parsing and class mapping.
We then showcase the two automatic annotations provided with our dataset.

The video demonstrates predictions from our six challenges. 
This showcases baseline results, but on a training sequence demonstrating `near perfect' performance as opposed to current baseline performance.
This aims to highlight the potential of \newDataset{} and the link between these challenges. Our Video demonstration is available at: \textcolor{blue}{\underline{\url{https://youtu.be/MUlyXDDzbZU}}}

\section{Further Collection Details}
\label{app:sectionB}

In this section we provide further details of how \newDataset{} was collected including comparing to the annotation pipeline from our previous work~\cite{Damen2018EPICKITCHENS}.

\chParagraph{Camera Settings for Collection}
Head mounted GoPro Hero 7 was used for data collection filming at 50fps with video stabilisation. 
Our choice of 50 fps avoids overhead light flickering visible in~\cite{Damen2018EPICKITCHENS} that occurs due to the difference between frame rates and the national grid frequency.

\begin{figure*}[t]
    \centering
    \includegraphics[width=0.9\textwidth]{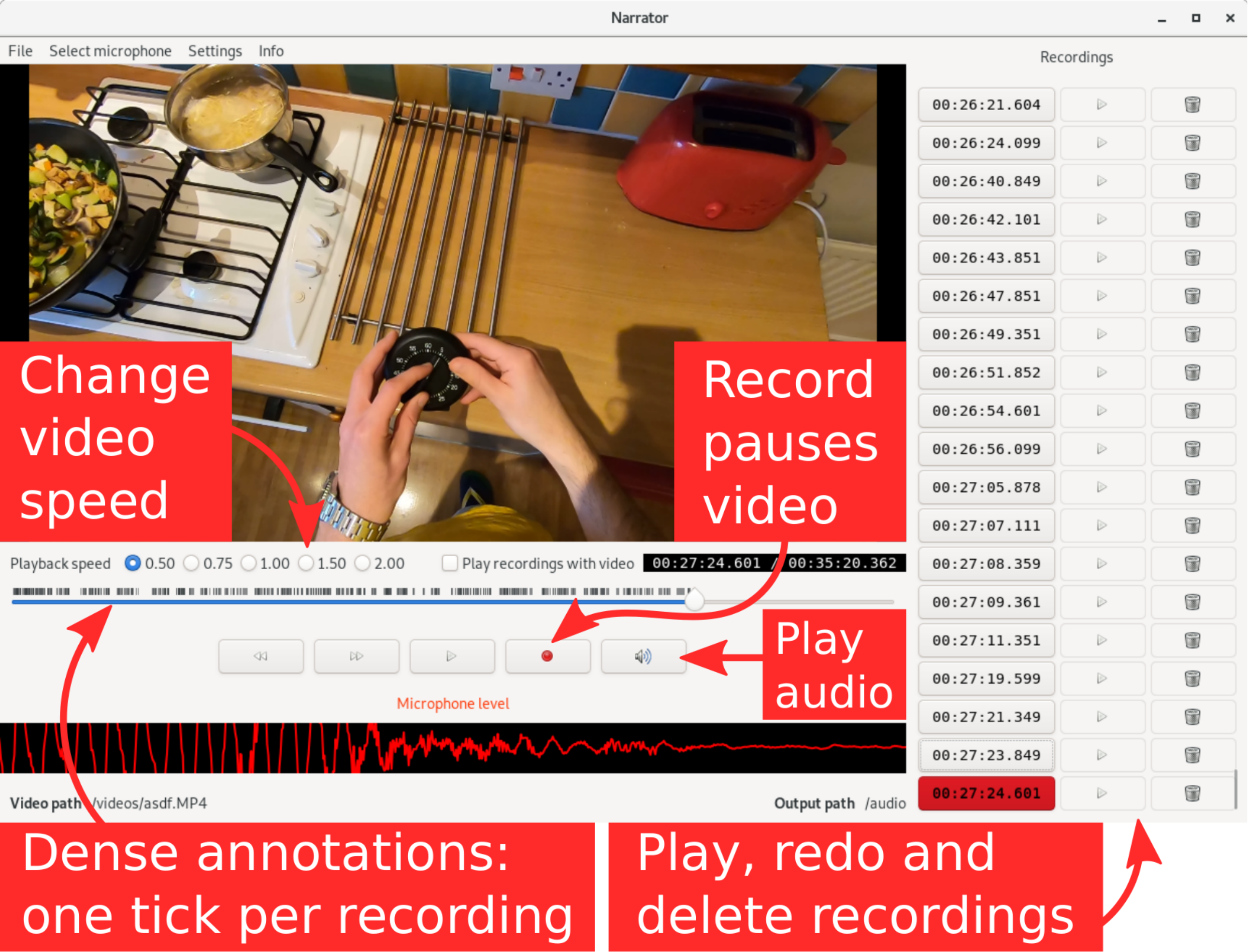}
    \caption{Components of our \annStyle{} annotation tool.}
    \label{fig:narrator}
\end{figure*}

\chParagraph{Narration \annStyle{} interface}
Fig.~\ref{fig:narrator} contains a more detailed look at our proposed \annStyle{} narrator.
Annotators had a number of options to help with the recording, including whether or not to hear the audio from the captured video while narrating, and the ability to change the speed of the video. They could also play, redo or delete recordings they had already made.

As mentioned in Section~2, this led to denser and more correct annotations, as annotators were able to pause the video while providing annotations, avoiding any missed annotations of critical actions.

\chParagraph{Transcription}
Thanks to our \annStyle{} narrator, each audio clip contained a single action narration, whereas formerly speech chunks were combined into 30 second clips.
In~\cite{Damen2018EPICKITCHENS}, Amazon Mechanical Turk (AMT) workers had to translate and transcribe this audio narration in a single step.
To ensure correctness and consistency, we split the transcription from the translation steps.
The set of non-English transcriptions was first agreed by multiple annotators and then translated in one go by a hired translator.

Additionally, we provided images during the transcription step centred around the timestamp collected by the \annStyle{} Narrator at $\{-0.25s, 0s, +0.25s\}$ to improve context (see Fig.~1b).

\chParagraph{Temporal Annotator}
Previously, initial start/end times were obtained by automatic alignment of captions using  YouTube automatic subtitling API. This is problematic as it assumes action length is the same as the narration length.
We adopt a different approach here starting from our accurate single timestamps produced by our proposed \annStyle{} narrator.
We developed a temporal segment annotation interface (see Fig 1d), where annotators start from this rough-time stamp and annotate the start/end time.
We also increased the number of annotators per segment to 5, compared to 4 used in~\cite{Damen2018EPICKITCHENS}. This resulted in higher agreements between annotators.

\section{Challenges' Implementation Details}
\label{sec:app_impl_details}

In this section we include the implementation and training details for all of the baselines, to enable replication of our results. Additionally, for some challenges, further details are provided such as definition of evaluation metrics.

\subsection{Action Recognition}
\label{subsec:challenge_action_recog_supp}

\chParagraph{Implementation and Training Details} 
We use our publicly available PyTorch~\cite{pytorch} model definitions of TSN~\cite{wang2016tsn}, TRN~\cite{zhou2017trn} and TSM~\cite{lin2019tsm}. 
We use ResNet-50 backbones for all models with publicly available initialisations - these are ImageNet weights for TSN and TRN and Kinetics weights for TSM.
We train two instances of each model: one with 8 RGB frames as input, and the other with 8 stacks of 5 $(u, v)$ flow fields computed using TV-$L_1$~\cite{zach2007duality}.
We use two-way output in the last layer, one to predict verbs and the other to predict nouns with an average verb/noun loss.
Actions are predicted as the most likely verb-noun combinations computed by combining softmaxed verb/noun scores.

We train each model for 80 epochs using SGD with momentum 0.9 and a learning rate of 0.01 decayed at epochs 20 and 40 by a factor of 10.
TSN and TRN models are trained on 8 GPUs with a batch-size of 128, whereas TSM used a batch-size of 64 on 4 GPUs.
We apply a weight decay of 0.0005 to all weights in the models, drop out with $p=0.7$, and clipping gradients above 20.
We use center-crop evaluation.
The RGB and optical flow models are trained individually, and predictions are averaged pre-softmax during inference.

For TBN, we use the publicly available PyTorch~\cite{pytorch} model from~\cite{kazakos2019epic}. We train using a batch size of 64, 6 segments, and drop the learning rate at epoch 40 and 60. All unspecified hyperparameters remain unchanged. 

For SlowFast~\cite{feichtenhofer2019slowfast}, we use the publicly available PyTorch~\cite{pytorch} model. We modify the model to have a two-way output for verbs and nouns, and train it with the average verb-noun loss. We use the SlowFast 8x8, ResNet-50 backbone, initialised from Kinetics pretrained weights also provided by \cite{feichtenhofer2019slowfast}. A 1~second clip randomly sampled from the video is used as input to the model during training. We train for 30 epochs using SGD with momentum 0.9 and a learning rate of 0.01 decayed at epochs 20 and 25 by a factor of 10. 
The model is trained on 8 GPUs with a batch-size of 32, using a weight decay of 0.0001 to all weights in the model and drop out with $p=0.5$. We freeze all batch-normalisation layers' parameters and statistics during training. During testing, we uniformly sample 10 clips ($1s$ each) from each video, and a single center crop per clip, and average their predictions. 

\subsection{Weakly-Supervised Action Recognition}

\chParagraph{Implementation and Training Details}
We use our publicly available PyTorch~\cite{pytorch} code from~\cite{Moltisanti_2019_CVPR} for both baselines.
This uses TSN~\cite{wang2016tsn} with Inception backbone and batch normalisation~\cite{ioffe2015batch}, pre-trained on Kinetics-400~\cite{carreira2017quo}. Predictions employ standard late-fused two-stream approach at test time (RGB and Flow models are trained independently). This uses 25 RGB frames (or optical flow stacks) for testing. 

We set a length of 5 seconds for the fixed-length segment baseline. For this baseline, frames are sampled randomly from equally sized segments (as proposed in~\cite{wang2016tsn}). For the baseline from~\cite{Moltisanti_2019_CVPR} training frames are selected using the sampling distributions which are iteratively updated. For both baselines we sample 5 frames for training. The ADAM~\cite{kingma2014adam} optimiser is used with initial learning rate equal to 0.0001 halved twice during training, and report results after 80 epochs. We changed the parameters from~\cite{Moltisanti_2019_CVPR} as follows: $w=2.5$ seconds and $s=0.75$, updating the distributions every 5 epochs with $(\lambda_c, \lambda_w, \lambda_s) = (0.5, 0.25, 0.25)$. We set CL $h = 1$ and CL $z = 0.25$.
Update proposals are generated with $\tau \in \{0.5, 0.85\}$, discarding proposals with length less than 10 frames.  

\subsection{Action Detection}
\chParagraph{Implementation and Training Details}
We train Boundary Matching Network (BMN)~\cite{lin2019bmn} using the publicly available implementation\footnote{\url{https://github.com/JJBOY/BMN-Boundary-Matching-Network}} to produce temporal action proposals. BMN is trained using TSN-based features, as in action recognition.
As proposed in~\cite{lin2019bmn}, we rescale the feature sequence of each video to the length of the observation window $l_\omega$. Since the proposed dataset contains videos of different lengths, we choose a large observation window $l_\omega=400$ and set the maximum action length to $D=400$. To limit the amount of memory required at training time, we set the number of sample points to $N=4$. We train one model on the Train set for $9$ epochs, which maximizes performance on Val. We use this model to report on both Val and Test. We apply Soft Non-Maximum Suppression with the parameters suggested in~\cite{lin2019bmn} to reduce the number of overlapping proposals and retain the top scoring $1,000$ instances per video. 

Each proposal is then classified using the SlowFast Network with implementation details as in Section~\ref{subsec:challenge_action_recog_supp}. Note that we classify proposals on the validation set using the SlowFast model trained only on the training set, whereas we classify proposals on the test set using the model trained on the union of the training and validation sets.

\subsection{Action Anticipation}

\chParagraph{Implementation and Training Details}
We follow our prior work~\cite{furnari2019would} training a TSN model to extract RGB and Flow features, using the same hyperparameters recommended in~\cite{furnari2019would}.
The RGB model has been trained for $95$ epochs, while the optical flow branch has been trained for $132$ epochs, which maximise performance on Val.
Object-based features are extracted running the object detector from~\cite{furnari2019would}, trained on manually-annotated object bounding boxes from our previous edition~\cite{Damen2018EPICKITCHENS}. 
The RU-LSTM model is trained using the provided implementation with SGD and a fixed learning rate of $0.01$. The single-modality RGB, optical flow and object branches are pre-trained with sequence completion respectively for $88$, $95$, and $98$ epochs, then fine-tuned for the anticipation task for $86$, $81$ and $7$ epochs respectively. The full architecture with modality attention is trained for $29$ epochs. These maximise performance on Val. All other parameters are kept as their default values in the public code from~\cite{furnari2019would}, The same model is used to report both on Val and Test. 

\chParagraph{\edits{Impact of current action on anticipation}}
\edits{Predicting a future action given the currently observed one provides a strong prior. To assess this, we created three co-occurrence matrices for verbs, nouns and actions. Each matrix $M$ is constructed such that $M[i,j]$ reports the number of times class $j$ is observed after class~$i$ in the training set considering $\tau_a=1$ as the anticipation time. At test time, we rely on the last observed action $i$ to predict the most frequent $5$ actions following $i$ (corresponding to the $5$ largest values of the $i^{th}$ row of $M$). 
Note that this calculation requires knowledge of the observed action from the ground truth, thus cannot be considered a baseline, as it cannot be replicated in inference.
We found that this oracle knowledge of the current action
obtains $20.84\%$, $25.00\%$ and $8.92\%$ for Top-5 verb, noun and action labels respectively on the validation set. These numbers are significantly larger than the chance baseline ($6.39\%$, $2.00\%$, $0.20\%$) from Table~\ref{tab:aa_results} but still lower than the ones of the RU-LSTM baseline ($27.76\%$, $30.76\%$, $14.04\%$). These results suggest that, while the prior is indeed a strong one, as you would hope for meaningful sequences of actions, the considered baseline is going beyond recognising the current action and applying an action sequence prior.}

\subsection{Unsupervised Domain Adaptation (UDA) for Action Recognition}

\chParagraph{Validation Splits for Hyper-parameter Tuning}
As the target domain is unlabelled, no labelled data is available for hyper-parameter tuning. Therefore, we split the training data to define a \textit{Source Val} and \textit{Target Val} splits with data collected by 4 of the 16 participants. Of these, 2 participants are of returning kitchens and 2 of changing kitchens. The \textit{Source Train} and \textit{Target Train} are thus composed of the 12 remaining participants. 

For hyper-parameter tuning, models are trained on labelled data from \textit{Source Val} and unlabelled from \textit{Target Val}. The performance on \textit{Target Val} can be used to asses the impact of different hyper-parameters. 

To obtain the results for the leaderboard and accompanying challenge, a new model is trained on \textit{Source Train} and unlabelled \textit{Target Train}, using the hyper-parameters optimised from the validation split. This model is evaluated on \textit{Target Test} to obtain results.

\chParagraph{Note on zero-shot actions}
Due to the unscripted nature of the data collection, a negligible number of verb and noun classes in the target domain are not present in the source domain, $0.2\%$ and $2.3\%$ respectively. We have not removed these to maintain the same splits used in other challenges. Additionally, $9.46\%$ actions (exact verb-noun combinations) did not exist in the targets domain, these are referred to as the zero-shot actions. Note it is still possible to predict these actions as both verbs and nouns were present in the source domain.

\chParagraph{Implementation and Training Details}
We train the TBN feature extractor on the union of \textit{Source Train} and \textit{Source Val}. We make these features publicly available.
We use the available code from~\cite{Chen2019}, to train and evaluate `Source-Only' as well as `TA3N' baselines.
 We modify the code to consider multi-modal input, by concatenating the features from all modalities as input. This automatically increased the number of parameters in the first fully connected layer.
    
    We improve the performance of TA3N by initialising the domain discriminators before the gradients are reversed and back-propagated. In our implementation, the domain discriminators' hyper-parameters are annealed similar to that in~\cite{ganin2016domain}:
    \begin{equation}
        \eta = \frac{2}{1+ exp(-p)} -1
        \label{eq:annealing}
    \end{equation}
    where $p$ is the training progress that linearly increases from 0 to 1.
    The domain discriminator hyperparameters are annealed up to the value specified in TA3N, i.e. $\lambda^s = 0.75\eta$, $\lambda^r=0.5\eta$ and $\lambda^t=0.75\eta$.
    The weighting of the categorical entropy on the target domain is set to $\gamma=0.003$.
    Models are trained for 30 epochs at a learning rate of $3e^{-3}$ reduced by a factor of 10 at epochs 10 and 20.

\subsection{Multi-Instance Action Retrieval}
\label{sec:appendix-retrieval}
\chParagraph{Evaluation Metrics}
We define the Relevance $\mathcal{R}$ between a video, $x_i$, and a caption, $c_j$, as given by the averaged Intersection-over-Union of the verb and noun classes:

\begin{equation}
    \mathcal{R}(x_i, c_j) = \frac{1}{2} \left(\frac{|x_i^v \cap c_j^v|}{|x_i^v \cup c_j^v|} + \frac{|x_i^N \cap c_j^N|}{|x_i^N \cup c_j^N|}\right)
    \label{eq:retrieval-relevance}
\end{equation}
where $x_i^v$ is the set of verb classes in the video and $c_j^N$ is the set of noun classes in the caption.

The nDCG can be calculated for a query video, $x_i$, and the ranked list of gallery captions, $C_r$, as the Discounted Cumulative Gain (DCG) over the Ideal Discounted Cumulative Gain (IDCG):
\begin{equation}
    nDCG(x_i, C_r) = \frac{DCG(x_i, C_r)}{IDCG(x_i, C_r)}
\end{equation}
with the DCG being given by:
\begin{equation}
    DCG(x_i, C_r) = \sum_{j=1}^{|C_r|} \frac{\mathcal{R}(x_i, c_j)}{log(j+1)}
    \label{eq:ndcg}
\end{equation}
\edits{To calculate the $IDCG(x_i, C_r)$, we need the ground truth ranking between video $x_i$ and captions $C_r$.
    To do this, we first find the relevance between video $x_i$ and every caption in $C_r$ as follows: $\{\mathcal{R}(x_i, c_j); \; \forall c_j \in C_r)\}$. We then construct $\hat{C}_r$, the ground truth ranking of captions, by sorting these in descending order of relevance. Note that if $\mathcal{R}(x_i, c_j) = \mathcal{R}(x_i, c_k)$ then $c_j$ and $c_k$ are ordered based on their unique ID due to the stable sort used, and similarly for the method to be evaluated.
    Finally, the $IDCG$ is calculated using} $IDCG(x_i, C_r)=DCG(x_i, \hat{C_r}))$.

$nDCG$ can be similarly defined for a query caption, $c_i$ and a gallery set of videos $X_r$.

\chParagraph{Implementation and Training Details}
For video features we use 25 RGB, Flow and Audio features extracted uniformly from TBN~\cite{kazakos2019epic}.
We make these features publicly available.
Features from each modality are temporally averaged and then concatenated to provide the final feature vector for each video, with size 3072.
Text features come from word2vec~\cite{mikolov2013efficient} trained on the wikipedia corpus with an embedding space of size 100.

The MLP baseline uses a 2 layer perceptron which projects both the visual and textual features into the same embedding space.
We set the final embedding size to 512 and the size of the hidden units is 1280 and 78 for visual/textual respectively (halfway between initial feature size and output space size).
MLP is trained for 100 epochs with a batch size of 64 and a learning rate of $0.01$.
Triplets are sampled randomly using the semantic relevance used when calculating mAP/nDCG (\emph{i.e.} verb and noun class are identical), with triplets being sampled every 10 iterations.
The triplet loss terms for all four pairs of modalities are set to 1.0, apart from the the text-to-visual weight which is assigned a weight of 2.0.

We use our public code of JPoSE~\cite{wray2019fine} . 
Each Part-of-Speech embedding is modelled off of the MLP baseline, but using the part-of-speech relevancies defined in~\cite{wray2019fine} (e.g.~for the verb embedding the verb class between two captions must be the same).
The final embeddings are concatenated and fed into a final fully connected layer with shared weights for the action embedding.
The verb and noun embedding spaces have an output embedding size of 256, with the resulting action embedding space having an output size of 512.
Triplets are independently resampled (randomly) every 10 epochs.
A batch size of 64 is used with a learning rate of $0.01$ and the model is trained for 100 epochs.

\end{document}